\documentclass[runningheads]{llncs}

\usepackage{eccv}

\usepackage{eccvabbrv}

\usepackage{booktabs}
\usepackage{adjustbox}
\usepackage{subcaption}
\usepackage{caption}
\usepackage{graphicx}
\usepackage{tikz}
\usepackage{makecell}
\usepackage{nicematrix}
\usepackage{array}
\usepackage{multirow}
\usepackage{algorithm}
\usepackage{algpseudocode}
\usepackage{float}
\usepackage{comment}

\newbox{\bigpicturebox}

\newcommand{\quotes}[1]{``#1''}
\newcommand{\minisection}[1]{\vspace{0.02in} \noindent {\bf #1}\ \ }

\usepackage[accsupp]{axessibility}  

\usepackage{hyperref}

\usepackage{orcidlink}

\begin{document}
\sloppy

\title{Enhancing Perceptual Quality in Video Super-Resolution through Temporally-Consistent Detail Synthesis using Diffusion Models} 

\titlerunning{Stable Video Super-Resolution (StableVSR)}

\author{Claudio Rota\inst{1}\orcidlink{0000-0002-6086-9838} \and
Marco Buzzelli\inst{1}\orcidlink{0000-0003-1138-3345} \and
Joost van de Weijer\inst{2}\orcidlink{0000-0002-9656-9706}}

\authorrunning{C.~Rota et al.}

\institute{
University of Milano-Bicocca, Milan, Italy\\
\email{\{claudio.rota, marco.buzzelli\}@unimib.it}\\ \and
Universitat Aut\`{o}noma de Barcelona, Barcelona, Spain\\
\email{joost@cvc.uab.es}}

\maketitle
\begin{center}
    \centering
    \renewcommand{\arraystretch}{0.1}
    \captionsetup{type=figure}
    \adjustbox{width=\textwidth}{
    \begin{tabular}{ccccc}
        Reference frame&BasicVSR$++$ & RVRT & StableVSR (ours) & Reference\\
        \includegraphics[height=0.25\textwidth]{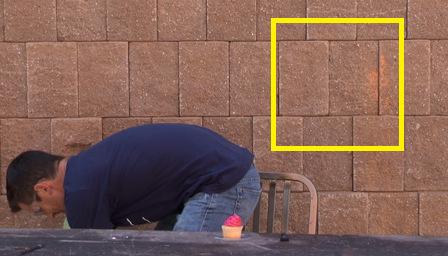} & 
        \includegraphics[width=0.25\textwidth]{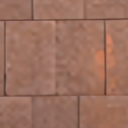} &
        \includegraphics[width=0.25\textwidth]{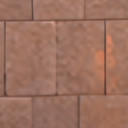} &
        \includegraphics[width=0.25\textwidth]{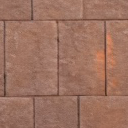} & \includegraphics[width=0.25\textwidth]{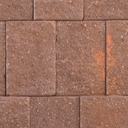}\\
        & \makecell{29.61 / 0.383} & \makecell{\textbf{29.64} / 0.379} & \makecell{27.88 \textbf{0.191}}\\
         
        \includegraphics[height=0.25\textwidth] {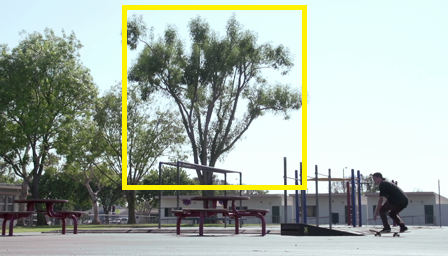} & 
        \includegraphics[width=0.25\textwidth]{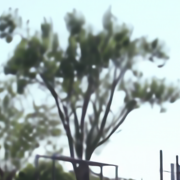} &
        \includegraphics[width=0.25\textwidth]{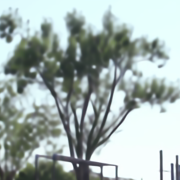} &
        \includegraphics[width=0.25\textwidth]{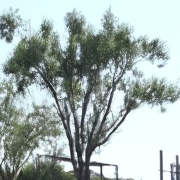} & \includegraphics[width=0.25\textwidth]{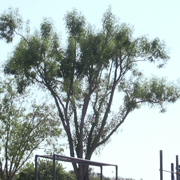}\\
        & \makecell{23.76 / 0.263} & \makecell{\textbf{24.11} / 0.260} & \makecell{21.20 /\textbf{0.105}}\\

        \includegraphics[height=0.25\textwidth]{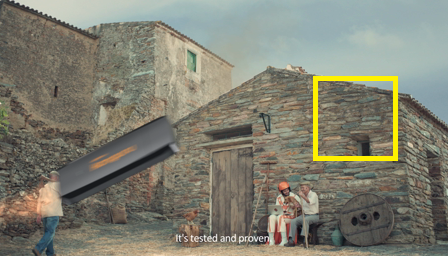} & 
        \includegraphics[width=0.25\textwidth]{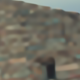} &
        \includegraphics[width=0.25\textwidth]{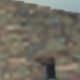} &
        \includegraphics[width=0.25\textwidth]{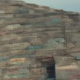} &
        \includegraphics[width=0.25\textwidth]{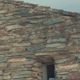}\\
        & \makecell{26.82 / 0.407} &  \makecell{\textbf{26.87} / 0.408} & \makecell{25.14 / \textbf{0.196}}\\

    \end{tabular}
    }
    \captionof{figure}{Reconstruction metrics, such as PSNR, evaluate the pixel-wise difference and do not correlate well with human perception. Perceptual metrics, such as LPIPS, better capture the perceptual quality. Existing methods lack generative capability and focus on reconstruction quality, often producing perceptually unsatisfying results. The proposed StableVSR enhances the perceptual quality by synthesizing realistic details, leading to better visual results. Results reported as PSNR / LPIPS using $\times4$ upscaling. Best results in bold text. PSNR: the higher, the better. LPIPS: the lower, the better.}
    \label{fig:intro}
\end{center}

\begin{abstract}
    In this paper, we address the problem of enhancing perceptual quality in video super-resolution (VSR) using Diffusion Models (DMs) while ensuring temporal consistency among frames. We present StableVSR, a VSR method based on DMs that can significantly enhance the perceptual quality of upscaled videos by synthesizing realistic and temporally-consistent details. We introduce the Temporal Conditioning Module (TCM) into a pre-trained DM for single image super-resolution to turn it into a VSR method. TCM uses the novel Temporal Texture Guidance, which provides it with spatially-aligned and detail-rich texture information synthesized in adjacent frames. This guides the generative process of the current frame toward high-quality and temporally-consistent results.
    In addition, we introduce the novel Frame-wise Bidirectional Sampling strategy to encourage the use of information from past to future and vice-versa. This strategy improves the perceptual quality of the results and the temporal consistency across frames. We demonstrate the effectiveness of StableVSR in enhancing the perceptual quality of upscaled videos while achieving better temporal consistency compared to existing state-of-the-art methods for VSR. The project page is available at \url{https://github.com/claudiom4sir/StableVSR}.
  \keywords{Video super-resolution \and Perceptual quality \and Temporal consistency \and Diffusion models}
\end{abstract}

\section{Introduction}
\label{sec:intro}

Video super-resolution (VSR) aims to increase the spatial resolution of a video enhancing its level of detail and clarity.
Recently, many VSR methods based on deep learning techniques have been proposed~\cite{liu2022video}. Ideally, a VSR method should generate plausible new contents that are not present in the low-resolution frames. However, existing VSR methods lack generative capability and cannot synthesize realistic details. According to the perception-distortion trade-off, under limited model capacity, improving reconstruction quality inevitably leads to a decrease in perceptual quality~\cite{blau2018perception}. Existing VSR methods mainly focus on reconstruction quality. As a consequence, they often produce perceptually unsatisfying results~\cite{ledig2017photo}. As shown in Figure~\ref{fig:intro}, frames upscaled with recent state-of-the-art VSR
methods~\cite{chan2022basicvsr++,liang2022recurrent} have high reconstruction quality but low perceptual quality, exhibiting blurriness and lack of details~\cite{wang2018recovering}.

Diffusion Models (DMs)~\cite{ho2020denoising} are a class of generative models that transform random noise into images through an iterative refinement process. Inspired by the success of DMs in generating high-quality images~\cite{ho2020denoising,dhariwal2021diffusion,saharia2022photorealistic,rombach2022high}, several works have been recently proposed to address the problem of single image super-resolution (SISR) using DMs~\cite{li2022srdiff,saharia2022image,ho2022cascaded,gao2023implicit,sahak2023denoising,wang2024exploiting}. They show the effectiveness of DMs in synthesizing realistic textures and details, contributing to enhancing the perceptual quality of the upscaled images~\cite{ledig2017photo}. 
Compared to SISR, VSR requires the integration of information from multiple closely related but misaligned frames to obtain temporal consistency over time. Unfortunately, applying a SISR method to individual video frames may lead to suboptimal results
and may introduce temporal inconsistency~\cite{rota2023video}. Different approaches to encourage temporal consistency in video generation using DMs have been recently studied~\cite{blattmann2023align,wu2023tune,yu2023video,esser2023structure}. 
However, these methods do not specifically address VSR and
do not use fine-texture temporal guidance. As a consequence, they may fail to achieve temporal consistency at fine-detail level, essential in the context of VSR.

In this paper, we address these problems and present \emph{Stable Video Super-Resolution} (StableVSR), a novel method for VSR based on DMs. StableVSR enhances the perceptual quality of upscaled videos by synthesizing
realistic and temporally-consistent details.
StableVSR exploits a pre-trained DM for SISR~\cite{rombach2022high} to perform VSR by introducing the \emph{Temporal Conditioning Module} (TCM). TCM guides the generative process of the current frame toward the generation of high-quality and temporally-consistent results over time. This is achieved by using the novel \emph{Temporal Texture Guidance}, which provides TCM with spatially-aligned and detail-rich texture information from adjacent frames: at every sampling step $t$, the predictions of the adjacent frames are projected to their initial state, \ie $t=0$, and spatially aligned to the current frame. 
At inference time, StableVSR uses the novel \emph{Frame-wise Bidirectional Sampling strategy} to avoid error accumulation problems and balance information propagation: a sampling step is first taken on all frames before advancing in sampling time, and information is alternately propagated forward and backward in video time.

In summary, our main contributions are the following:
\begin{itemize}
    \item We present \emph{Stable Video Super-Resolution} (StableVSR): the first work that approaches VSR under a generative paradigm using DMs. It 
    significantly enhances the perceptual quality of upscaled videos while ensuring temporal consistency among frames;
    \item We design the \emph{Temporal Texture Guidance} containing detail-rich and spatially-aligned texture information synthesized in adjacent frames. It guides the generative process of the current frame toward the generation of detailed and temporally consistent frames;
    \item We introduce the \emph{Frame-wise Bidirectional Sampling strategy} with forward and backward information propagation. It balances information propagation across frames and alleviates the problem of error accumulation; 
    \item We quantitatively and qualitatively demonstrate that the proposed StableVSR can achieve superior perceptual quality and better temporal consistency compared to existing methods for VSR.
\end{itemize}

\section{Related work}
\label{sec:relatedwork}

\minisection{Video super-resolution.}
Video super-resolution based on deep learning has witnessed considerable advances in the past few
years~\cite{liu2022video}. ToFlow~\cite{xue2019video} fine-tuned a pre-trained optical flow estimation network with the rest of the framework to achieve more accurate frame alignment. TDAN~\cite{tian2020tdan} proposed the use of deformable convolutions~\cite{zhu2019deformable} for spatial alignment as an alternative to optical flow computation. EDVR~\cite{wang2019edvr} extended the alignment module proposed in TDAN~\cite{tian2020tdan} to better handle large motion and used temporal attention~\cite{vaswani2017attention} to balance the contribution of each frame.
BasicVSR~\cite{chan2021basicvsr} revised the essential components for a VSR method, \ie bidirectional information propagation and spatial feature alignment, and proposed a simple yet effective solution. BasicVSR$++$~\cite{chan2022basicvsr++} improved BasicVSR~\cite{chan2021basicvsr} by adding second-order grid propagation and flow-guided deformable alignment. RVRT~\cite{liang2022recurrent} combined recurrent networks with the attention mechanism~\cite{vaswani2017attention} to better capture long-range frame dependencies and enable parallel frame predictions. RealBasicVSR~\cite{chan2022investigating} proposed to use a pre-cleaning module before applying a variant of BasicVSR~\cite{chan2021basicvsr}, and the use of a discriminator model~\cite{wang2021realesrgan} to improve the perceptual quality of the results.

\minisection{Diffusion Models for single image super-resolution.} The success of Diffusion Models in image generation~\cite{ho2020denoising,dhariwal2021diffusion,saharia2022photorealistic,rombach2022high} inspired the development of single image super-resolution methods based on DMs~\cite{li2022srdiff,saharia2022image,ho2022cascaded,gao2023implicit,sahak2023denoising,wang2024exploiting}. SRDiff~\cite{li2022srdiff} and SR3~\cite{saharia2022image} demonstrated DMs can achieve impressive results in SISR. SR3+~\cite{sahak2023denoising} extended SR3~\cite{saharia2022image} to images in the wild by proposing a higher-order degradation scheme and noise conditioning augmentation. LDM~\cite{rombach2022high} proposed to work in a VAE latent space~\cite{esser2021taming} to reduce complexity requirements and training time. CDM~\cite{ho2022cascaded} proposed to cascade multiple DMs to achieve SISR at arbitrary scales.
IDM~\cite{gao2023implicit} proposed to introduce the implicit image function in the decoding part of a DM to achieve continuous super-resolution. StableSR~\cite{wang2024exploiting} leveraged prior knowledge encapsulated in a pre-trained text-to-image DM to perform SISR avoiding intensive training from scratch.

\section{Background on Diffusion Models}
\label{sec:background}
Diffusion Models~\cite{ho2020denoising} convert a complex data distribution $x_0 \sim p_{data}$ into a simple Gaussian distribution $x_T \sim \mathcal{N}(0, I)$, and then recover data from it. A DM is composed of two processes: the diffusion process and the reverse process.

\minisection{Diffusion process.} The diffusion process is a Markov chain that corrupts data $x_0 \sim p_{data}$ until they approach Gaussian noise $x_T \sim \mathcal{N}(0, I)$ after $T$ diffusion steps. It is defined as:
\begin{equation}
    q(x_1, ..., x_T | x_0) = \prod_{t=1}^{T} q(x_t|x_{t-1}) \; ,
\end{equation}
where $t$ represents a diffusion step and $q(x_t | x_{t-1}) = \mathcal{N}(x_t; \sqrt{1 - \beta_t}(x_{t-1}),\beta_t I)$, with $\beta_t$ being a fixed or learnable variance schedule. At any step $t$, $x_t$ can be directly sampled from $x_0$ as:
\begin{equation}
\label{eq:xt}
    x_t = \sqrt{\overline{\alpha}_t}x_0 + \sqrt{1 - \overline{\alpha}_t}\epsilon \; ,
\end{equation}
where $\alpha_t = 1 - \beta_t$, $\overline{\alpha}_t = \prod_{i=1}^t \alpha_i$ and $\epsilon \sim \mathcal{N}(0, I)$. 

\minisection{Reverse process.} The reverse process is a Markov chain that removes noise from $x_T \sim \mathcal{N}(0, I)$ until data $x_0 \sim p_{data}$ are obtained. It is defined as:
\begin{equation}
    p_\theta(x_0, ..., x_{T-1}|x_T) = \prod_{t=1}^{T}p_\theta(x_{t-1}|x_t) \; ,
\end{equation}
where $p_\theta(x_{t-1}|x_t) = \mathcal{N}(x_{t-1}; \mu_\theta(x_t, t), \Sigma_\theta I)$. The variance $\Sigma_\theta$ can be a learnable parameter~\cite{nichol2021improved} or a time-dependent constant~\cite{ho2020denoising}.
A neural network $\epsilon_\theta$ is trained to predict $\epsilon$ from $x_t$, and it can be used to estimate $\mu_\theta(x_t, t)$ as:
\begin{equation}
    \mu_\theta(x_t, t) = \frac{1}{\sqrt{\overline{\alpha}_t}} \left(x_t - \frac{1 - \alpha_t}{\sqrt{1 - \overline{\alpha}_t}}\epsilon_\theta(x_t, t)\right) \; .
\end{equation}
As a consequence, we can sample $x_{t-1} \sim p_\theta(x_{t-1}|x_t)$ as:
\begin{equation}
\label{eq:prevt}
    x_{t-1} = \frac{1}{\sqrt{\alpha_t}} \left(x_t - \frac{1 - \alpha_t}{\sqrt{1 - \overline{\alpha}_t}}\epsilon_\theta(x_t, t)\right) + \sigma_tz \; ,
\end{equation}
where $z \sim \mathcal{N}(0, I)$ and $\sigma_t$ is the variance schedule.
In practice, according to Eq.~\ref{eq:xt}, we can directly predict $\tilde{x}_0$ from $x_t$ via projection to the initial state $t=0$ as:
\begin{equation}
\label{eq:x0}
    \tilde{x}_0 = \frac{1}{\sqrt{\overline{\alpha}_t}}\left(x_t - \sqrt{1 - \overline{\alpha}_t}\epsilon_\theta(x_t, t)\right) \; .
\end{equation}

\section{Methodology}
\label{sec:method}

\begin{figure}[t]
    \centering
    \includegraphics[width=\textwidth]{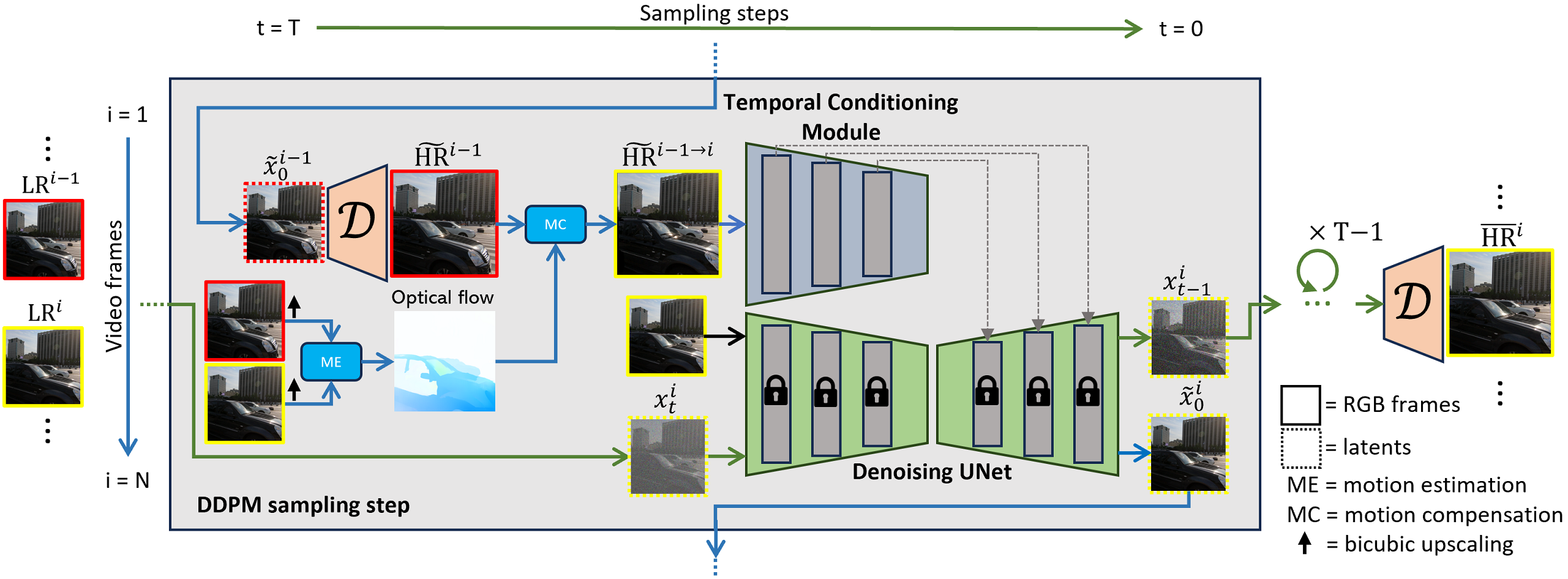}
    \caption{Overview of the proposed StableVSR. 
    We use the Temporal Conditioning Module (Section~\ref{subsec:tcm}) to turn a single image super-resolution LDM (denoising UNet) into a video super-resolution method. TCM exploits the novel Temporal Texture Guidance (Section~\ref{subsec:ttg}), which provides TCM with spatially-aligned and detail-rich texture information synthesized in adjacent frames.
    The sampling step is taken using the novel Frame-wise Bidirectional Sampling strategy (Section~\ref{subsec:bss}). $\mathcal{D}$ represents the VAE decoder. Green lines refer to progression in sampling time, while blue lines refer to progression in video time.}
    \label{fig:method}
\end{figure}
We present Stable Video Super-Resolution (StableVSR), a method for video super-resolution based on Latent Diffusion Models (LDMs)~\cite{rombach2022high}. StableVSR enhances the perceptual quality in VSR through temporally-consistent detail synthesis. 
The overview of the method is shown in Figure~\ref{fig:method}. Given a sequence of $N$ low-resolution frames $\{\text{LR}\}^N_{i=1}$, the goal is to obtain the upscaled sequence $\{\overline{\text{HR}}\}^N_{i=1}$.
StableVSR is built upon a pre-trained LDM for single image super-resolution~\cite{rombach2022high}, which is turned into a VSR method through the design and the addition of the Temporal Conditioning Module. TCM uses detail and structure information synthesized in adjacent frames to guide the generative process of the current frame. It allows obtaining high-quality and temporally-consistent frames over time. 
We design the Temporal Texture Guidance to provide TCM with rich texture information about the adjacent frames: at every sampling step, their predictions are projected to their initial state via Eq.~\ref{eq:x0}, converted into RGB frames, and aligned with the current frame via optical flow estimation and motion compensation. 
We introduce in StableVSR the Frame-wise Bidirectional Sampling strategy, where a sampling step is taken on all frames before advancing in sampling time, and information is alternately propagated forward and backward in video time. This alleviates the problem of error accumulation and balances the information propagation over time. A brief description of the pre-trained LDM for SISR~\cite{rombach2022high} is provided in the supplementary material.

\subsection{Temporal Conditioning Module}
\label{subsec:tcm}

Applying the SISR LDM~\cite{rombach2022high} to individual video frames introduces temporal inconsistency, as each frame is generated only based on the content of a single low-resolution frame.
In addition, this approach does not exploit the content shared among multiple video frames, leading to suboptimal results~\cite{rota2023video}.
We address these problems by introducing the Temporal Conditioning Module into the SISR LDM~\cite{rombach2022high}. The goal is twofold: (1) enabling the use of spatio-temporal information from multiple frames, improving the overall frame quality; (2) enforcing temporal consistency across frames. We use the information generated by the SISR LDM~\cite{rombach2022high} in the adjacent frames to guide the generative process of the current frame. In addition to obtaining temporal consistency, this solution provides additional sources of information to handle very small or occluded objects. TCM injects temporal conditioning into the decoder of the denoising UNet, as proposed in ControlNet~\cite{zhang2023adding}.

\subsection{Temporal Texture Guidance}
\label{subsec:ttg}

The Temporal Texture Guidance provides TCM with the texture information synthesized in adjacent frames. The goal is to guide the generative process of the current frame toward the generation of high-quality and temporally-consistent results.

\begin{figure}[t]
    \centering
    \adjustbox{width=.75\textwidth}{
    \setlength{\tabcolsep}{0.8pt}
    \renewcommand{\arraystretch}{0.4}
    \begin{NiceTabular}{ccccc}
         & $x_t$ & $\tilde{x}_0$ &  $||x_0- \tilde{x}_0||$ &  $||$HR $- \tilde{x}_0||$\\
         
        \raisebox{0.35\height}{\rotatebox{90}{$t=900$}} & 
        \includegraphics[width=0.15\textwidth]{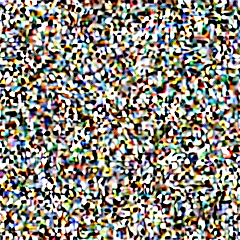} & \includegraphics[width=0.15\textwidth]{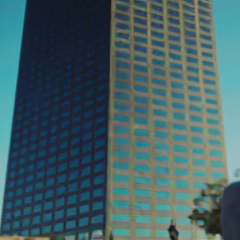} & \includegraphics[width=0.15\textwidth]{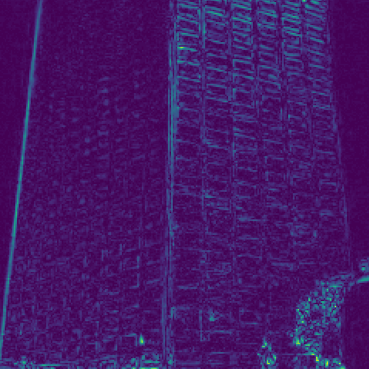}&
        \includegraphics[width=0.15\textwidth]{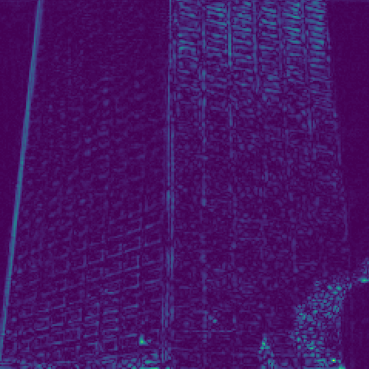}\\
        
        \raisebox{0.35\height}{\rotatebox{90}{$t=500$}} & 
        \includegraphics[width=0.15\textwidth]{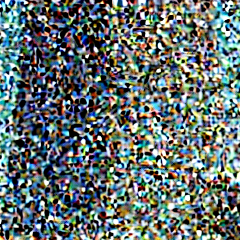} & \includegraphics[width=0.15\textwidth]{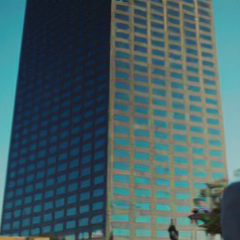} & \includegraphics[width=0.15\textwidth]{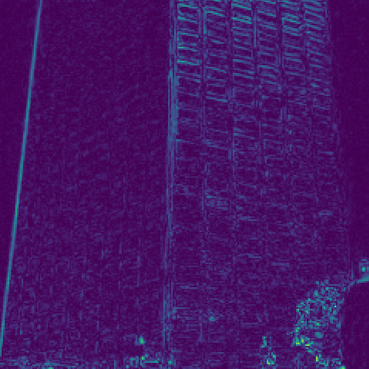}&
        \includegraphics[width=0.15\textwidth]{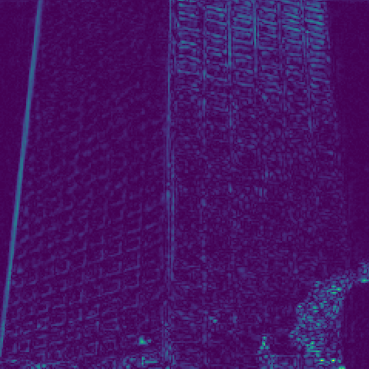}\\
        
        \raisebox{0.6\height}{\rotatebox{90}{$t=25$}} & 
        \includegraphics[width=0.15\textwidth]{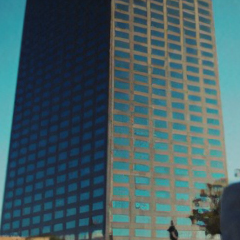} & \includegraphics[width=0.15\textwidth]{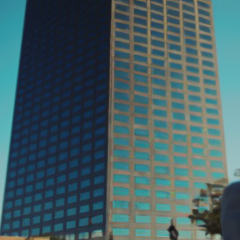} & \includegraphics[width=0.15\textwidth]{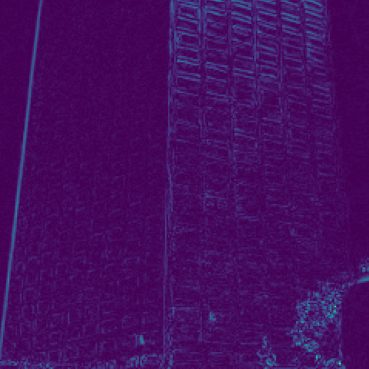}&
        \includegraphics[width=0.15\textwidth]{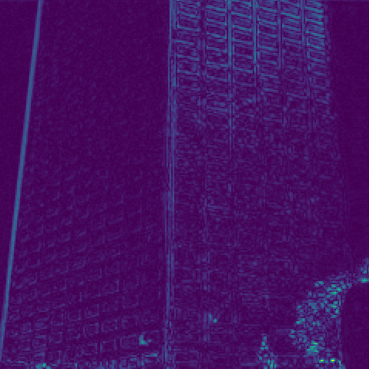}\\
    \end{NiceTabular}
    }
    \caption{Comparison between guidance on $x_t$ and $\tilde{x}_0$. Compared to $x_t$ (first column), $\tilde{x}_0$ computed via Eq.~\ref{eq:x0} contains very little noise regardless of the sampling step $t$ (second column). We can observe $\tilde{x}_0$ is closer to $x_0$ as $t$ decreases (third column). Here, $x_0$ corresponds to the last sampling step, \ie when $t=1$. In addition, $\tilde{x}_0$ increases its level of detail as $t$ decreases (fourth column).}
    \label{fig:x0examples}
\end{figure}

\minisection{Guidance on \boldmath{$\tilde{x}_0$}.} 
Using results of the previous sampling step $\{x_{t}\}^N_{i=1}$ as guidance to predict $\{x_{t-1}\}^N_{i=1}$, as proposed in~\cite{blattmann2023align,luo2023videofusion}, may not provide adequate texture information along the whole reverse process. This is because $x_t$ is corrupted by noise until $t$ approaches 0, as shown in Figure~\ref{fig:x0examples}. We address this problem by using a noise-free approximation of $x_t$, \ie $\tilde{x}_0$, to be used as guidance when taking a given sampling step $t$~\cite{fei2023generative}. This is achieved by projecting $x_t$ to its initial state, \ie $t=0$, using Eq~\ref{eq:x0}.
Since $\tilde{x}_0 \approx x_0$, it contains very little noise. In addition, it provides detail-rich texture information that is gradually refined as $t$ approaches 0, as shown in Figure~\ref{fig:x0examples}.

\minisection{Temporal conditioning.} We need to use information synthesized in adjacent frames to ensure temporal consistency. We achieve this by using $\tilde{x}_0$ obtained from the previous frame, \ie $\tilde{x}^{i-1}_0$, as guidance when generating the current frame. As $\tilde{x}^{i-1}_0$ is computed from $x^{i-1}_t$ using $\epsilon_\theta(x^{i-1}_t, t, \text{LR}^{i-1})$ via Eq.~\ref{eq:x0}, it contains the texture information synthesized in the previous frame at sampling step $t$.

\begin{figure}[t]
    \centering
    \setlength{\tabcolsep}{1pt}
    \adjustbox{width=0.7\textwidth}{
    \begin{tabular}{cc}
        \makecell{Motion compensation applied to $\tilde{x}_0$} & \makecell{Motion compensation applied to $\mathcal{D}(\tilde{x}_0)$}\\
        \includegraphics[width=0.45\columnwidth]{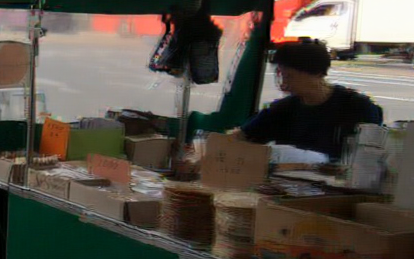} & \includegraphics[width=0.45\columnwidth]{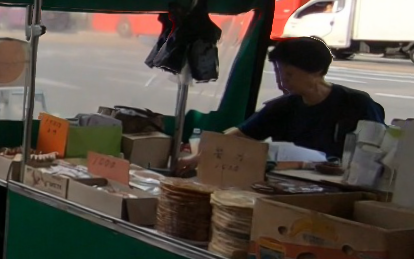}\\
    \end{tabular}
    }
    \caption{Comparison between applying motion compensation to $\tilde{x}_0$ in the latent space and to $\mathcal{D}(\tilde{x}_0)$ in the pixel domain. $\mathcal{D}$ represents the VAE decoder. In the first scenario, visible artifacts are introduced.}
    \label{fig:motioncompensation}
\end{figure}

\minisection{Spatial alignment.} Spatial alignment is essential to properly aggregate information from multiple frames~\cite{chan2021basicvsr}. The texture information contained in $\tilde{x}^{i-1}_0$ may not be spatially aligned with respect to the current frame due to video motion. We achieve spatial alignment via motion estimation and compensation, computing optical flow on the respective low-resolution frames $\text{LR}^{i-1}$ and $\text{LR}^i$. Directly applying motion compensation to $\tilde{x}^{i-1}_0$ in the latent space may introduce artifacts, as shown in Figure~\ref{fig:motioncompensation}. 
We address this problem by converting $\tilde{x}^{i-1}_0$ from the latent space to the pixel domain through the VAE decoder $\mathcal{D}$~\cite{esser2021taming} and then applying motion compensation.

\minisection{Formulation.} Given the previous and the current low-resolution frames $\text{LR}^{i-1}$ and $\text{LR}^i$, the current sampling step $t$ and the latent of the previous frame $x^{i-1}_t$, the Temporal Texture Guidance $\widetilde{\text{HR}}^{i-1\rightarrow i}$ is computed as:
\begin{equation}
\label{eq:guidance}
     \widetilde{\text{HR}}^{i-1\rightarrow i} = \text{MC(ME(LR}^{i-1}, \text{LR}^i), \mathcal{D}(\tilde{x}^{i-1}_0)) \; ,
\end{equation}
where MC is the motion compensation function, ME is the motion estimation method, $\mathcal{D}$ is the VAE decoder~\cite{esser2021taming} and $\tilde{x}^{i-1}_0$ is computed using $\epsilon_\theta(x^{i-1}_t, t, \text{LR}^{i-1})$ via Eq.~\ref{eq:x0}.

\subsection{Frame-wise Bidirectional Sampling strategy}
\label{subsec:bss}

Progressing all the sampling steps on one frame and using the result as guidance for the next frame in an auto-regressive manner, as proposed in~\cite{yu2023video}, may introduce the problem of error accumulation. In addition, unidirectional information propagation from past to future frames may lead to suboptimal results~\cite{chan2021basicvsr}. We address these problems by proposing the Frame-wise Bidirectional Sampling strategy:
we take a given sampling step $t$ on all the frames before taking the next sampling step $t-1$, alternately propagating information forward and backward in video time.
The pseudocode is detailed in Algorithm~\ref{alg:inference}. 
\begin{algorithm}[t]
\caption{Frame-wise Bidirectional Sampling strategy. ME and MC are \quotes{motion estimation} and \quotes{motion compensation}, respectively.}\label{alg:inference}
\begin{algorithmic}[1]
\scriptsize{
\Require Sequence of low-resolution frames $\{\text{LR}\}^N_{i=1}$; pre-trained $\epsilon_\theta$ for VSR, VAE decoder $\mathcal{D}$; method for ME.
\For{$i=1$ to $N$}
    \State $x^i_T = \mathcal{N}(0, I)$
\EndFor
\For{$t=T$ to 1}
    \For{$i=1$ to $N$}\Comment{Take sampling step $t$ on all the frames}
        \State $\widetilde{\text{HR}}^{i-1\rightarrow i} = \text{MC(ME(LR}^{i-1}, \text{LR}^i), \mathcal{D}(\tilde{x}^{i-1}_0))$ \textbf{if} $i > 1$ \Comment{Eq.~\ref{eq:guidance}}
        \State $\tilde{\epsilon}= \epsilon_\theta(x^i_t, t, \text{LR}^i, \widetilde{\text{HR}}^{i-1\rightarrow i})$ \textbf{if} $i > 1$ \textbf{else} $\epsilon_\theta(x^i_t, t, \text{LR}^i)$
        \State $z = \mathcal{N}(0, I)$ \textbf{if} $t> 1$ \textbf{else} 0
        \State $x^i_{t-1} = \frac{1}{\sqrt{\alpha}_t} \left(x^i_t - \frac{1 - \alpha_t}{\sqrt{1 - \overline{\alpha}_t}}\tilde{\epsilon}\right) + \sigma_tz$ \Comment{Eq.~\ref{eq:prevt}}
        \State $\tilde{x}^{i}_0 = \frac{1}{\sqrt{\overline{\alpha}_t}}\left(x^i_t - \sqrt{1 - \overline{\alpha}_t}\tilde{\epsilon}\right)$ \Comment{Eq.~\ref{eq:x0}}
    \EndFor
    \State Reverse sequence order of $\{x_{t-1}\}^N_{i=1}$, $\{\tilde{x}_0\}^N_{i=1}$ and $\{\text{LR}\}^N_{i=1}$
\EndFor\\
\Return $\{\overline{\text{HR}}\}^N_{i=1}\ = \{\mathcal{D}(x_0)\}^N_{i=1}$
}
\end{algorithmic}
\end{algorithm}
Given the latent $x^{i}_t$ at a sampling step $t$, the Temporal Texture Guidance $\widetilde{\text{HR}}^{i-1\rightarrow i}$ used by TCM is alternately computed via Eq.~\ref{eq:guidance} using $\tilde{x}^{i-1}_0$ or $\tilde{x}^{i+1}_0$, respectively related to the previous or the next frame. Information is propagated forward and backward in video time: the current frame is conditioned by past frames during forward propagation, and by future frames during backward propagation. Additional details are provided in the supplementary material. The first and the last frames of the sequence do not use TCM during forward and backward propagation, respectively. This is in line with other methods~\cite{chan2021basicvsr, chan2022basicvsr++}.

\subsection{Training procedure}
\label{subsec:training}

StableVSR is built upon a pre-trained LDM for SISR~\cite{rombach2022high}, hence we only need to train the Temporal Conditioning Module.  
\begin{algorithm}[t]
\caption{Training procedure. ME and MC are \quotes{motion estimation} and \quotes{motion compensation}, respectively.}\label{alg:training}
\begin{algorithmic}[1]
\scriptsize{
\Require Dataset $D$ with (LR, HR) pairs; pre-trained $\epsilon_\theta$ for SISR, method for ME.
\Repeat
    \State $(\text{LR}^{i-1}, \text{HR}^{i-1}), (\text{LR}^i,\text{HR}^i) \sim D$
    \State $x^{i-1}_0, x^{i}_0 = \mathcal{E}(\text{HR}^{i-1}), \mathcal{E}(\text{HR}^i)$
    \State $\epsilon^{i-1}, \epsilon^i \sim \mathcal{N}(0, I)$
    \State $t \sim \{0, ..., T\}$
    \State $\tilde{\epsilon}^{i-1} = \epsilon_\theta(\sqrt{\overline{\alpha}_t}x^{i-1}_0 + \sqrt{1 - \overline{\alpha}_t}\epsilon^{i-1}, t, \text{LR}^{i-1})$
    \State $\tilde{x}^{i-1}_0 = \frac{1}{\sqrt{\overline{\alpha}_t}}\left(x^i_t - \sqrt{1 - \overline{\alpha}_t}\tilde{\epsilon}^{i-1}\right)$ \Comment{Eq.~\ref{eq:x0}}
    \State $\widetilde{\text{HR}}^{i-1\rightarrow i} = \text{MC(ME(LR}^{i-1}, \text{LR}^i), \mathcal{D}(\tilde{x}^{i-1}_0))$ \Comment{Eq.~\ref{eq:guidance}}
    \State Take gradient descent step on:\\ 
    \hspace*{3em}$\nabla_\theta (||\epsilon^{i} - \epsilon_\theta(\sqrt{\overline{\alpha}_t}x^{i}_0 + \sqrt{1 - \overline{\alpha}_t}\epsilon^{i}, t,\text{LR}^i,\widetilde{\text{HR}}^{i-1\rightarrow i})||)$ 
\Until convergence
}
\end{algorithmic}
\end{algorithm}

We extend the ControlNet~\cite{zhang2023adding} training procedure by adding a step to compute the Temporal Texture Guidance $\widetilde{\text{HR}}^{i-1\rightarrow i}$ from the previous frame to be used for the current one. 
The pseudocode is detailed in Algorithm~\ref{alg:training}. Given two (LR, HR) pairs of consecutive frames (LR$^{i-1}$, HR$^{i-1}$) and (LR$^{i}$, HR$^i$), we first compute $x^{i-1}_0$ and $x^i_0$ by converting HR$^{i-1}$ and HR$^i$ into the latent space using the VAE encoder $\mathcal{E}$~\cite{esser2021taming}. We add $\epsilon \sim \mathcal{N}(0, I)$ to $x^{i-1}_0$ via Eq.~\ref{eq:xt}, obtaining $x^{i-1}_t$. We then compute $\tilde{x}^{i-1}_0$ using $x^{i-1}_t$ and $\epsilon_\theta(x^{i-1}_t, t,$ LR$^{i-1})$ via Eq.~\ref{eq:x0}, and we obtain $\widetilde{\text{HR}}^{i-1\rightarrow i}$ to be used for the current frame via Eq.~\ref{eq:guidance}. The training objective is:
\begin{equation}
 \mathbb{E}_{t,x^i_0,\epsilon,\text{LR}^i,\widetilde{\text{HR}}^{i-1\rightarrow i}}[||\epsilon - \epsilon_\theta(x^i_t, t, \text{LR}^i, \widetilde{\text{HR}}^{i-1\rightarrow i})||_2] \; ,
\end{equation}
where $t \sim [1, T]$ and $x^i_t$ is obtained by adding $\epsilon \sim \mathcal{N}(0, I)$ to $x^i_0$ via Eq.~\ref{eq:xt}.

\section{Experiments}
\label{sec:exp}

\subsection{Implementation details}
\label{subsec:impdetails}

StableVSR is built upon Stable Diffusion $\times4$ Upscaler\footnote{\url{https://huggingface.co/stabilityai/stable-diffusion-x4-upscaler}} (SD$\times4$Upscaler), which uses the low-resolution images as guidance via concatenation. 
SD$\times4$Upscaler uses a VAE decoder~\cite{esser2021taming} with $\times 4$ upscaling factor to perform super-resolution. We use the same decoder in our StableVSR. 
The architecture details are described in the supplementary material.
In all our experiments, the results are referred to $\times4$ super-resolution.
We add the Temporal Conditioning Module via ControlNet~\cite{zhang2023adding} and train it for 20000 steps. The training procedure is described in Algorithm~\ref{alg:training}. We use RAFT~\cite{teed2020raft} for optical flow computation.
We use 4 NVIDIA Quadro RTX 6000 for our experiments. We use the Adam optimizer~\cite{kingma2014adam} with a batch size set to 32 and the learning rate fixed to $1e-5$. 
Randomly cropped patches of size $256\times256$ with horizontal flip are used as data augmentation.
We use DDPM~\cite{ho2020denoising} sampling with $T=1000$ during training and $T=50$ during inference.

\subsection{Datasets and evaluation metrics}
\label{subsec:setup}

We adopt two benchmark datasets for the evaluation of the proposed StableVSR: Vimeo-90K~\cite{xue2019video} and REDS~\cite{nah2019ntire}. Vimeo-90K~\cite{xue2019video} contains 91701 7-frame video sequences at 448 $\times$ 256 resolution. It covers a broad range of actions and scenes. Among these sequences, 64612 are used for training and 7824 (called Vimeo-90K-T) for evaluation. REDS~\cite{nah2019ntire} is a realistic and dynamic scene dataset containing 300 video sequences. Each sequence has 100 frames at 1280 $\times$ 720 resolution. Following previous works~\cite{chan2021basicvsr, chan2022basicvsr++}, we use the sequences 000, 011, 015, and 020 (called REDS4) for evaluation and the others for training.

We evaluate perceptual quality using LPIPS~\cite{zhang2018unreasonable} and DISTS~\cite{ding2020image}. The results evaluated with additional perceptual metrics~\cite{ke2021musiq, wang2023exploring,mittal2012making} are reported in the supplementary material.
For temporal consistency evaluation, we adopt tLP~\cite{chu2020learning} and tOF~\cite{chu2020learning}, using RAFT~\cite{teed2020raft} for optical flow computation. 
We also report reconstruction metrics like PSNR and SSIM~\cite{wang2004image} for reference. 

\subsection{Comparison with state-of-the-art methods}
\label{subsec:sotacomparison}

\begin{table}[t]
\caption{Quantitative comparison with state-of-art methods for VSR. Perceptual metrics are marked with $\star$, reconstruction metrics with $\diamond$, and temporal consistency metrics with $\bullet$. Best results in bold text. All the perceptual metrics highlight the proposed StableVSR achieves better perceptual quality. Temporal consistency metrics show that StableVSR achieves better temporal consistency.}
\label{tab:results}
\centering
\adjustbox{width=\textwidth}{%
\begin{tabular}{ccccccccccccc}
\toprule
                 \multirow{2}{*}{VSR method}& \multicolumn{6}{c}{Vimeo-90K-T}                         & \multicolumn{6}{c}{REDS4}\\\cmidrule(lr){2-7}\cmidrule(lr){8-13}
 &
  tLP$\bullet$$\downarrow$ &
  tOF$\bullet$$\downarrow$ & 
  LPIPS$\star$$\downarrow$ &
  DISTS$\star$$\downarrow$&
  PSNR$\diamond$$\uparrow$ &
  SSIM$\diamond$$\uparrow$ &
  tLP$\bullet$$\downarrow$ &
  tOF$\bullet$$\downarrow$ &
  LPIPS$\star$$\downarrow$ &
  DISTS$\star$$\downarrow$&
  PSNR$\diamond$$\uparrow$ &
  SSIM$\diamond$$\uparrow$
   \\\midrule
Bicubic &
 12.47 & 2.23 & 0.289 & 0.209 & 29.75 & 0.848 &  22.72 & 4.04
 & 0.453 & 0.186 & 26.13 & 0.729  \\

ToFlow  & 4.96 & 1.53 & 0.152 &  0.150 & 32.28      &   0.898  & -     &   -   &  -     &   - & - & - \\

EDVR  & - & -        &     -  &    -   &  -     &   -   &  9.18 & 2.85 & 0.178 & 0.082 & 31.02 & 0.879 \\
TDAN & 4.89 & 1.50   &   0.120    &   0.122    &  34.10      & 0.919     &  -     & -    &    -   &   - & - & -     \\
MuCAN & 4.85 & 1.50 & 0.097 & 0.108 & 35.38 & 0.934 &  9.15 & 2.85 & 0.185 & 0.085 & 30.88 & 0.875 \\
BasicVSR   & 4.94 & 1.54 & 0.103 & 0.113 & 35.18 & 0.931 &  9.91 & 2.87 & 0.165 & 0.081 & 31.39 & 0.891  \\
BasicVSR$++$ & 4.35 & 1.75 & 0.092 & 0.105 & 35.69 & 0.937 &  9.02 & 2.75 & 0.131 & 0.068 & 32.38 & 0.907\\
RVRT   & 4.28 & 1.42 & 0.088 & 0.101 & \textbf{36.30} & \textbf{0.942} & 8.97 & 2.72 & 0.128 & 0.067 & \textbf{32.74} & \textbf{0.911} \\
RealBasicVSR & - & -& - & - & - & - & 6.44 & 4.74 & 0.134 & 0.060 & 27.07 & 0.778\\
StableVSR (ours) & \textbf{3.89} & \textbf{1.37} & \textbf{0.070} & \textbf{0.087} & 31.97 & 0.877 & \textbf{5.57} & \textbf{2.68} & \textbf{0.097} & \textbf{0.045} & 27.97 & 0.800
\\\bottomrule
\end{tabular}%
}
\end{table}

\begin{figure}[t]
    \centering
    \adjustbox{width=\textwidth}{
    \begin{tabular}{ccccc}
    Sequence 82, clip 798 (Vimeo-90K-T) & BasicVSR$++$ & RVRT & StableVSR (ours) & Reference\\
    \includegraphics[height=0.25\textwidth]{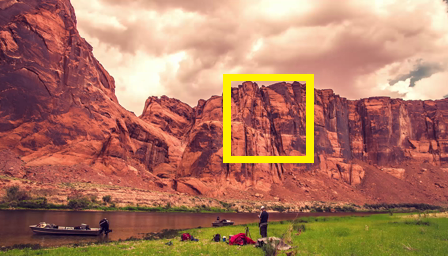} & \includegraphics[width=0.25\textwidth]{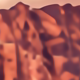} & \includegraphics[width=0.25\textwidth]{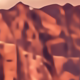} & 
    \includegraphics[width=0.25\textwidth]{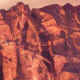} & 
    \includegraphics[width=0.25\textwidth]{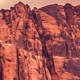}\\

    Clip 015, frame 38 (REDS4) & RVRT & RealBasicVSR & StableVSR (ours) & Reference\\
    \includegraphics[height=0.25\textwidth]{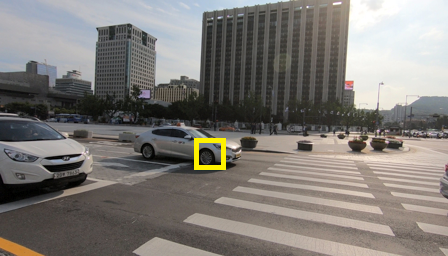} & \includegraphics[height=0.25\textwidth]{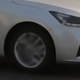} & \includegraphics[height=0.25\textwidth]{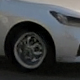} & 
    \includegraphics[height=0.25\textwidth]{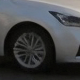} & \includegraphics[height=0.25\textwidth]{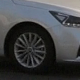}\\
    
    \end{tabular}}
    \caption{Qualitative comparison with state-of-the-art methods for VSR. The proposed StableVSR better enhances the perceptual quality of the upscaled frames by synthesizing more realistic details.}
    \label{fig:results}
\end{figure}

\begin{figure}[t]
    \centering
    \adjustbox{width=\textwidth}{
    \begin{tabular}{cc}
         \includegraphics[width=\textwidth]{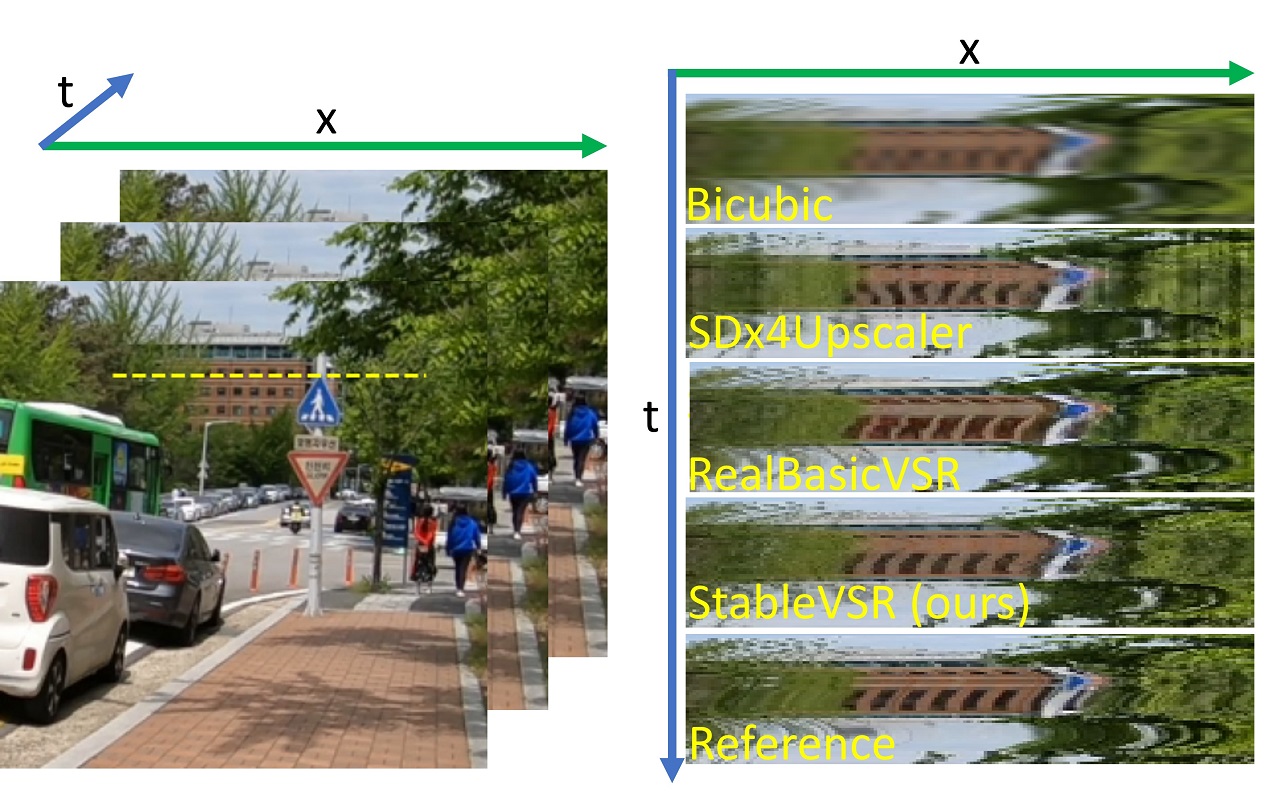}&\includegraphics[width=\textwidth]{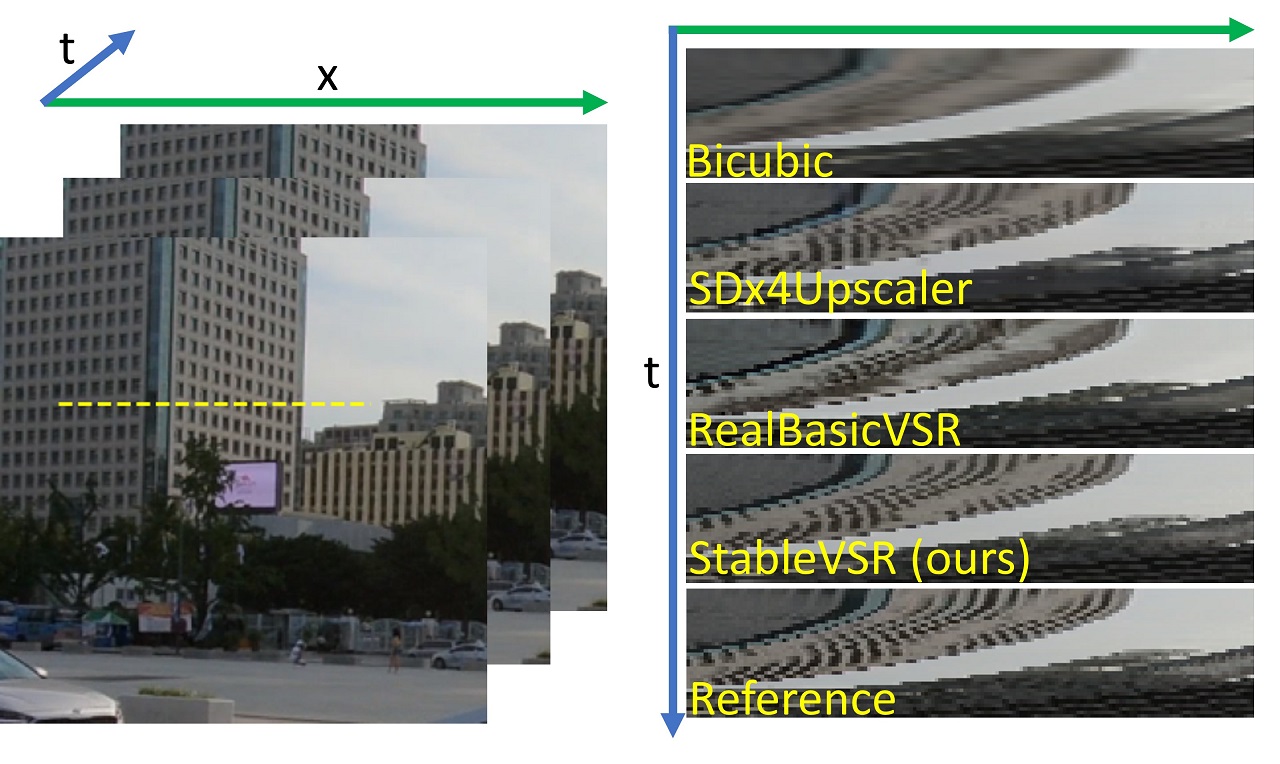}
    \end{tabular}}
    \caption{Comparison of temporal profiles. We consider a frame row and track the changes over time. The temporal profile of StableVSR is more regular than SD$\times4$Upscaler and more similar to the reference profiles than RealBasicVSR, reflecting a better consistency over time. Results on sequences 000 and 015 of REDS4, respectively.}
    \label{fig:temporalprofiles}
\end{figure}

We compare StableVSR with other state-of-the-art methods for VSR, including ToFlow~\cite{xue2019video}, EDVR~\cite{wang2019edvr},  TDAN~\cite{tian2020tdan},
MuCAN~\cite{li2020mucan},
BasicVSR~\cite{chan2021basicvsr}, BasicVSR$++$~\cite{chan2022basicvsr++}, RVRT~\cite{liang2022recurrent}, and RealBasicVSR~\cite{chan2022investigating}. Note that RealBasicVSR~\cite{chan2022investigating} is a generative method based on GANs~\cite{goodfellow2020generative}.
The quantitative comparison is reported in Table~\ref{tab:results}. 

\minisection{Frame quality results.}
As shown in Table~\ref{tab:results}, StableVSR outperforms the other methods in perceptual quality metrics. This is also confirmed by the qualitative results shown in Figure~\ref{fig:results}: the frames upscaled by StableVSR look more natural and realistic. Additional results are reported in the supplementary material. StableVSR and RealBasicVSR~\cite{chan2022investigating}, due to their generative nature, can synthesize details that cannot be found in the spatio-temporal frame neighborhood. This is because they capture the semantics of the scenes and synthesize missing information accordingly. Compared to RealBasicVSR~\cite{chan2022investigating}, StableVSR generates more natural and realistic details, leading to higher perceptual quality. In Table~\ref{tab:results}, we can observe StableVSR has poorer performance in PSNR and SSIM~\cite{wang2004image}. This is in line with the perception-distortion trade-off~\cite{blau2018perception}. Nevertheless, StableVSR achieves better reconstruction quality than bicubic upscaling and RealBasicVSR~\cite{chan2022investigating}. 

\begin{figure}[t]
    \centering
    \setlength{\tabcolsep}{2pt}
    \adjustbox{width=\textwidth}{
    \begin{tabular}{ccccc}
    SD$\times4$Upscaler & RVRT & RealBasicVSR & StableVSR (ours) & Reference\\

        \includegraphics[width=0.25\textwidth]{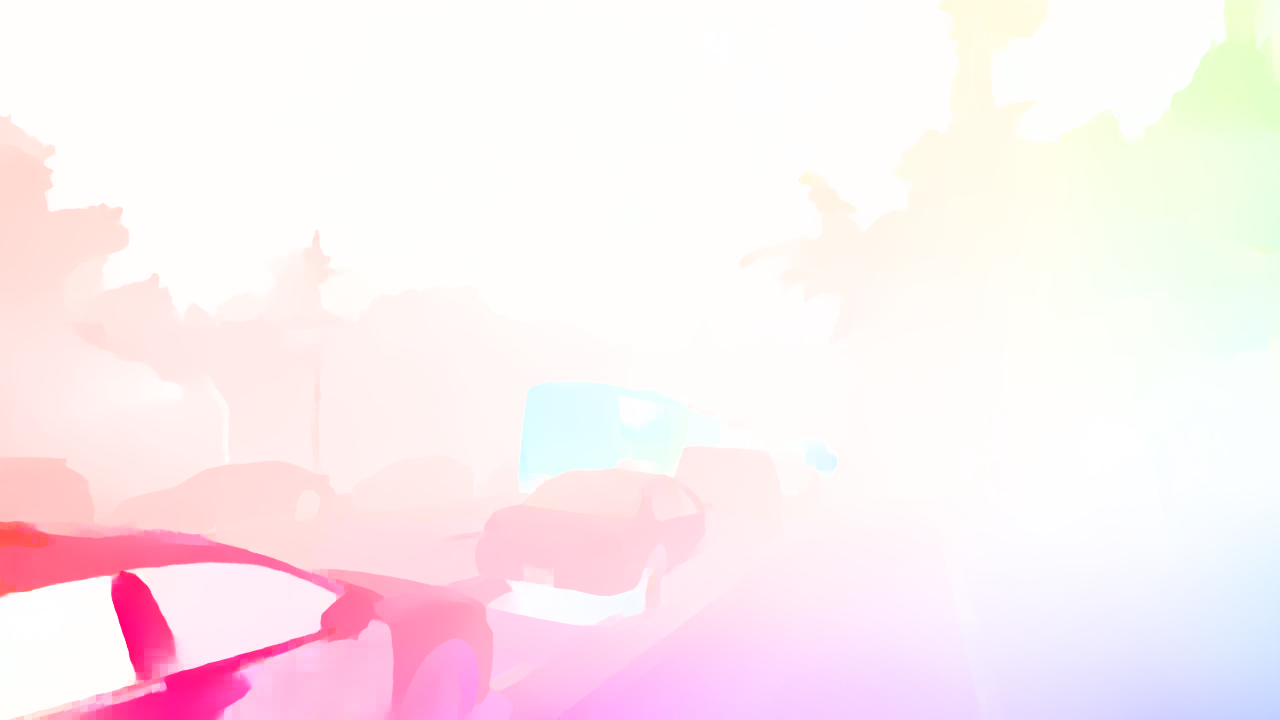} &
        \includegraphics[width=0.25\textwidth]{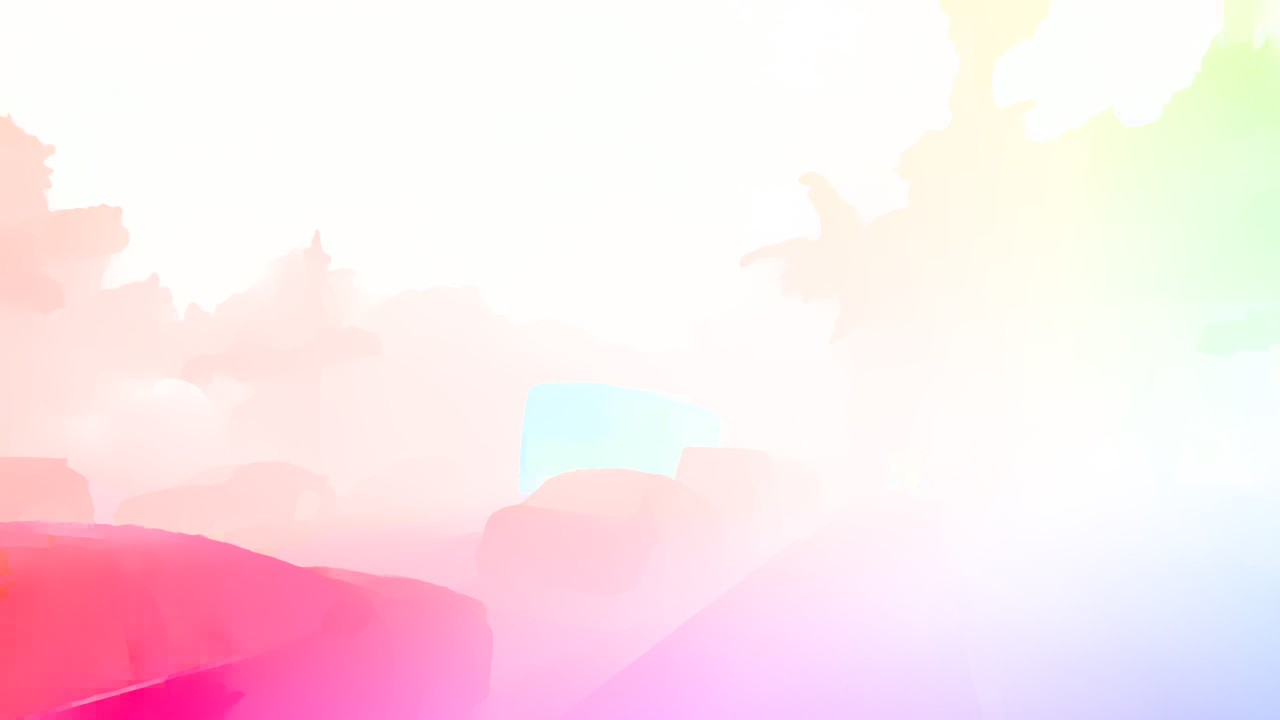} & \includegraphics[width=0.25\textwidth]{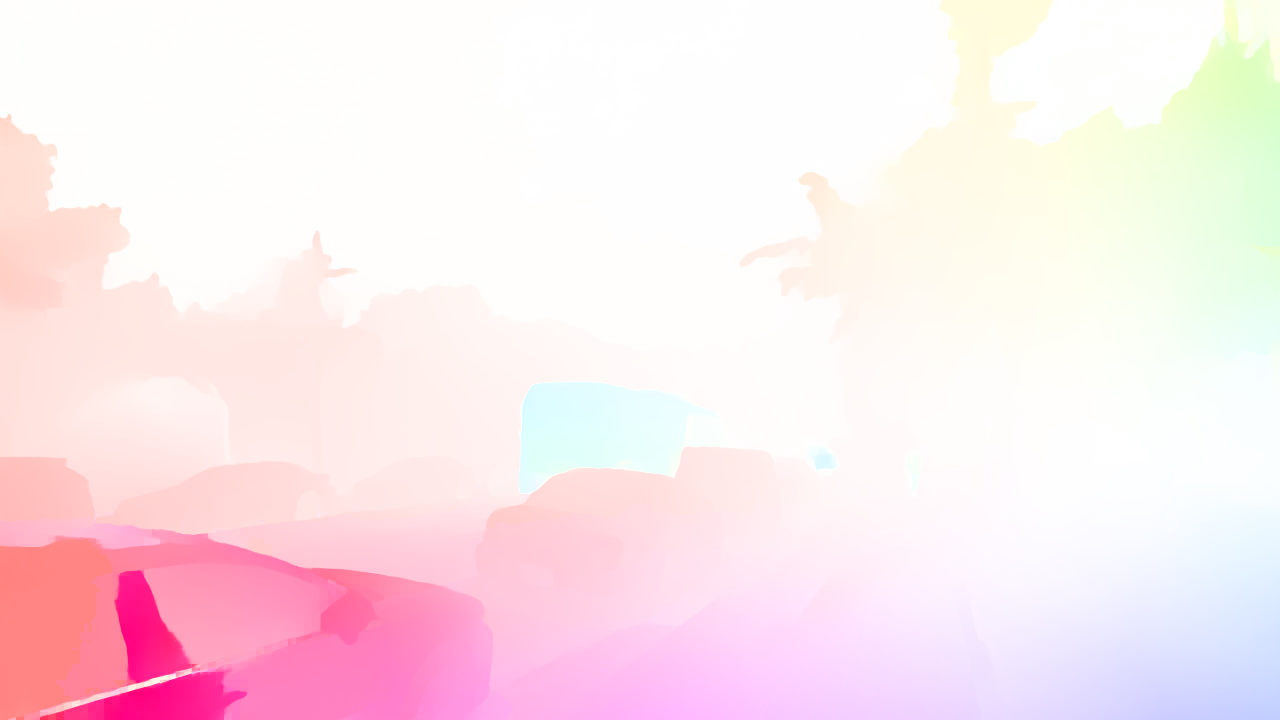} & \includegraphics[width=0.25\textwidth]{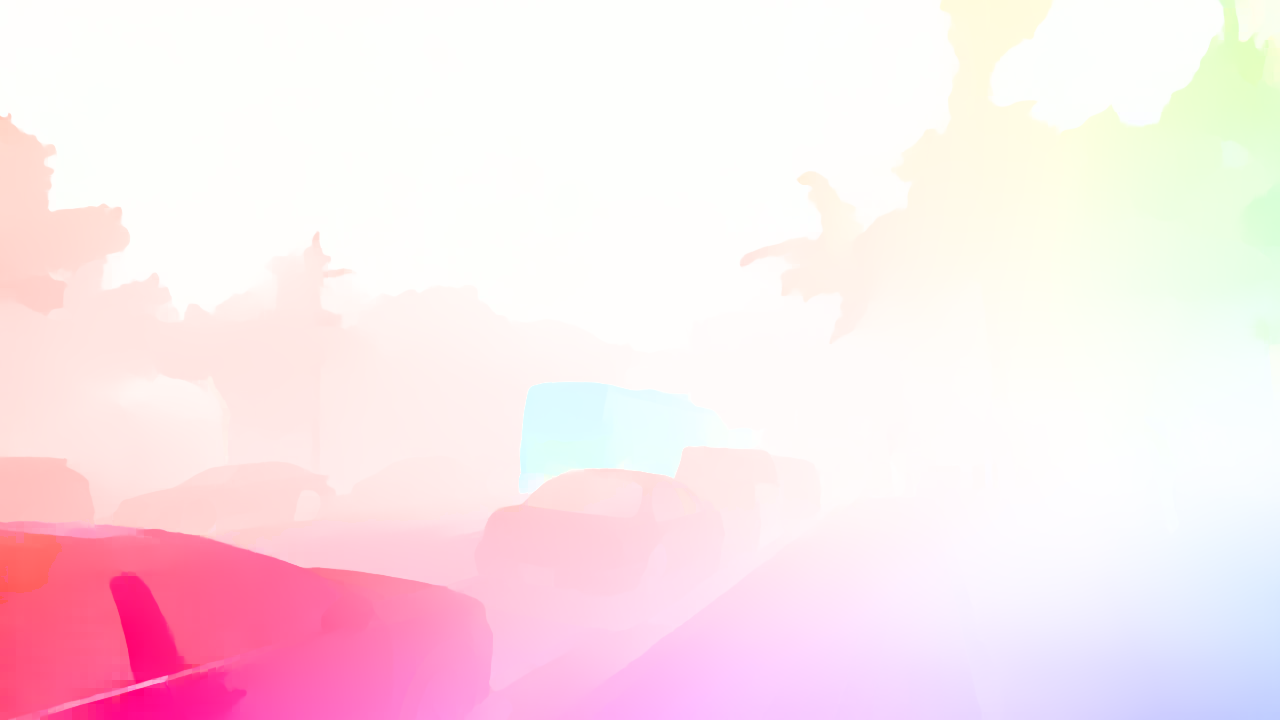} & \includegraphics[width=0.25\textwidth]{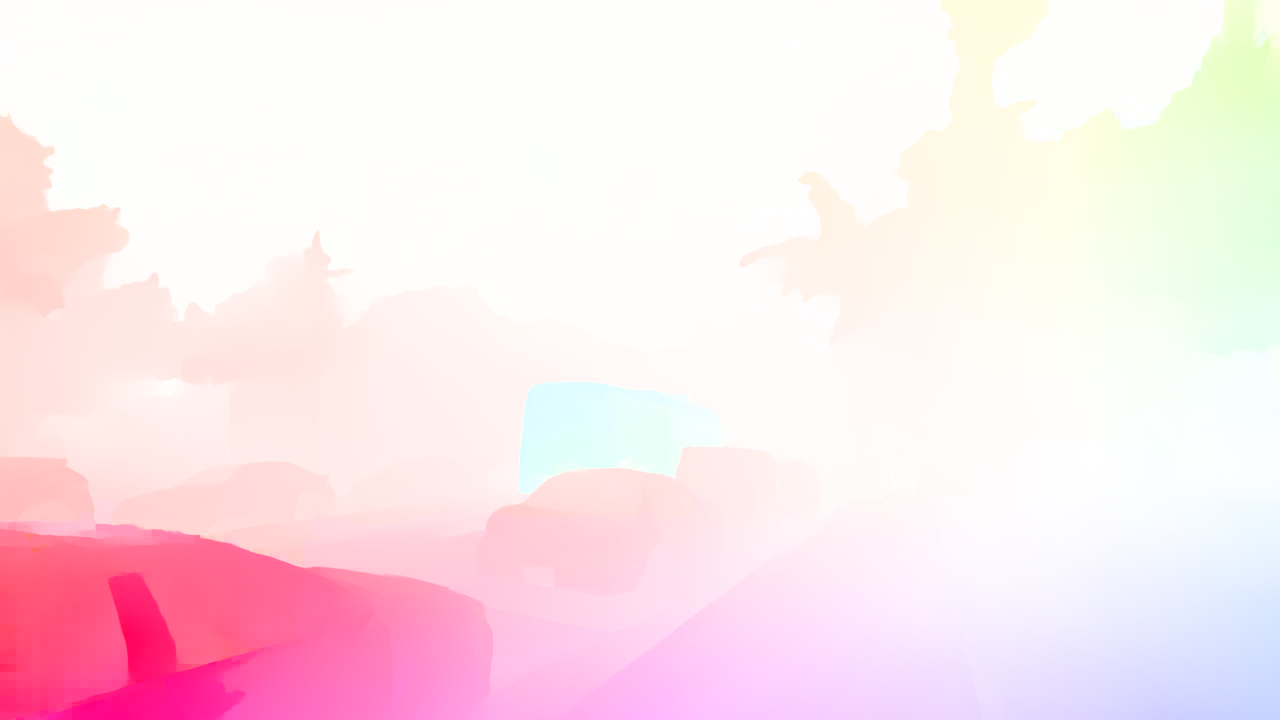}\\
        
        \includegraphics[width=0.25\textwidth]{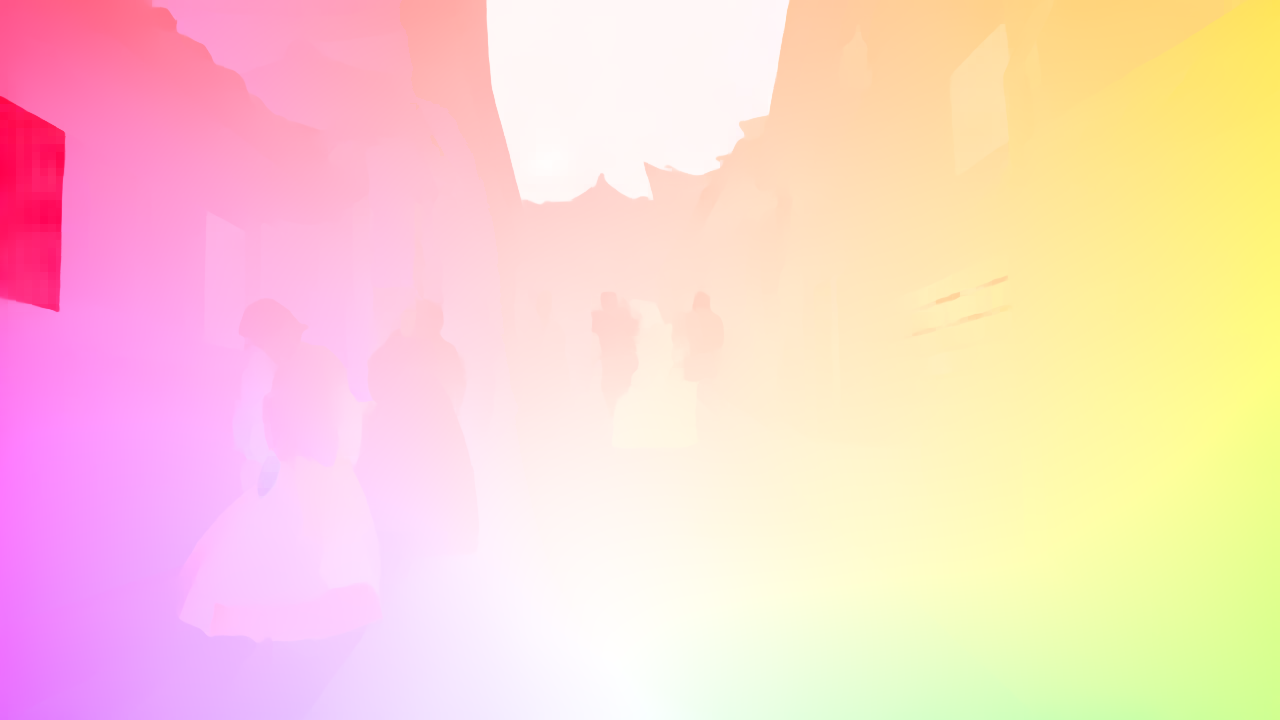} &
        \includegraphics[width=0.25\textwidth]{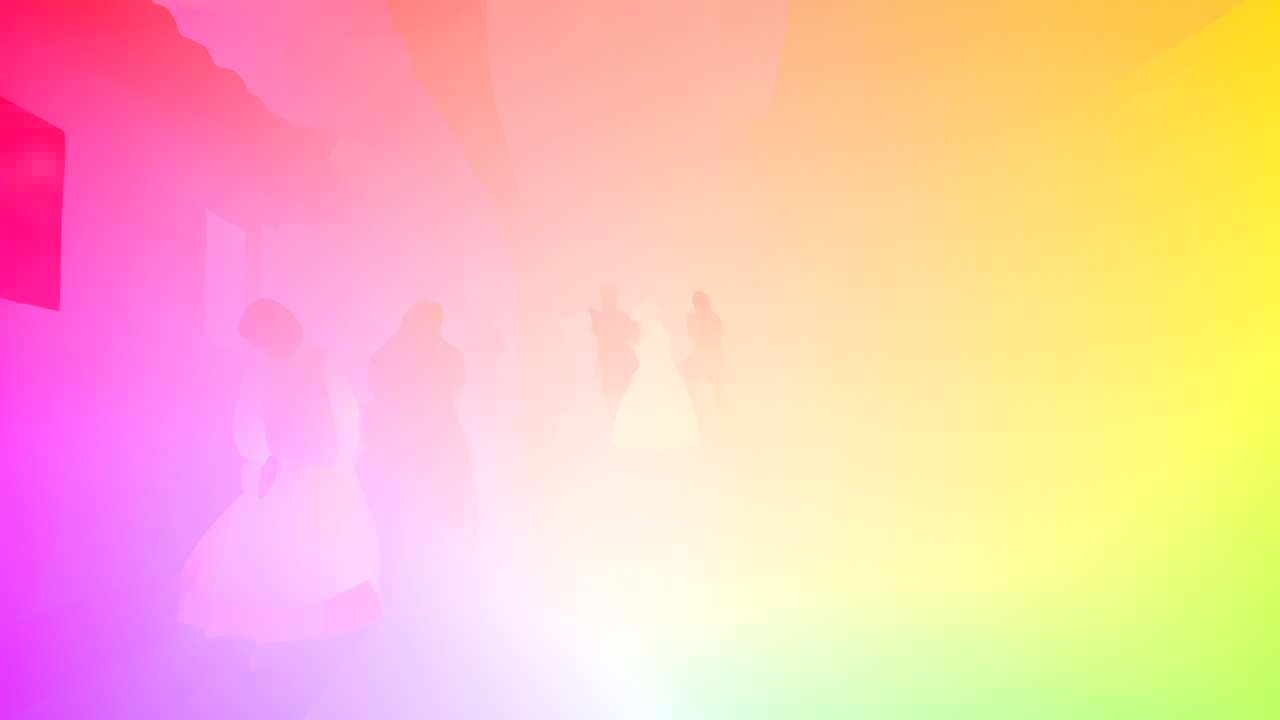} & \includegraphics[width=0.25\textwidth]{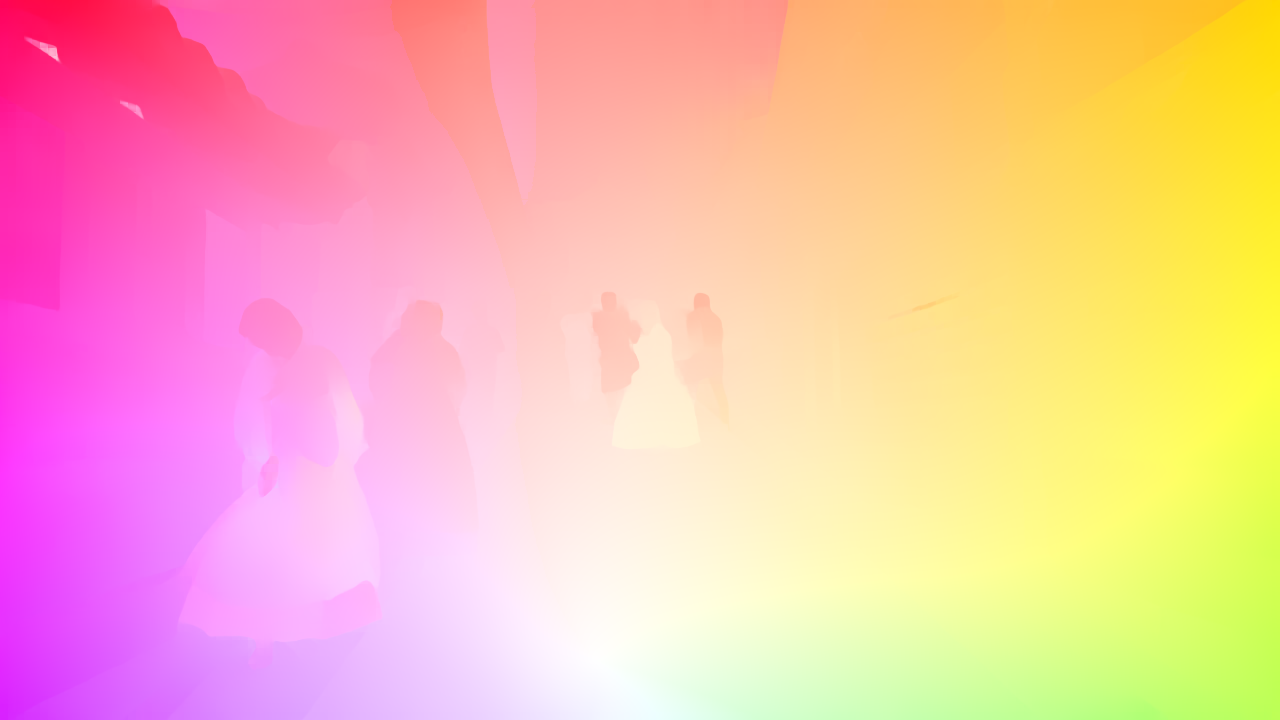} & \includegraphics[width=0.25\textwidth]{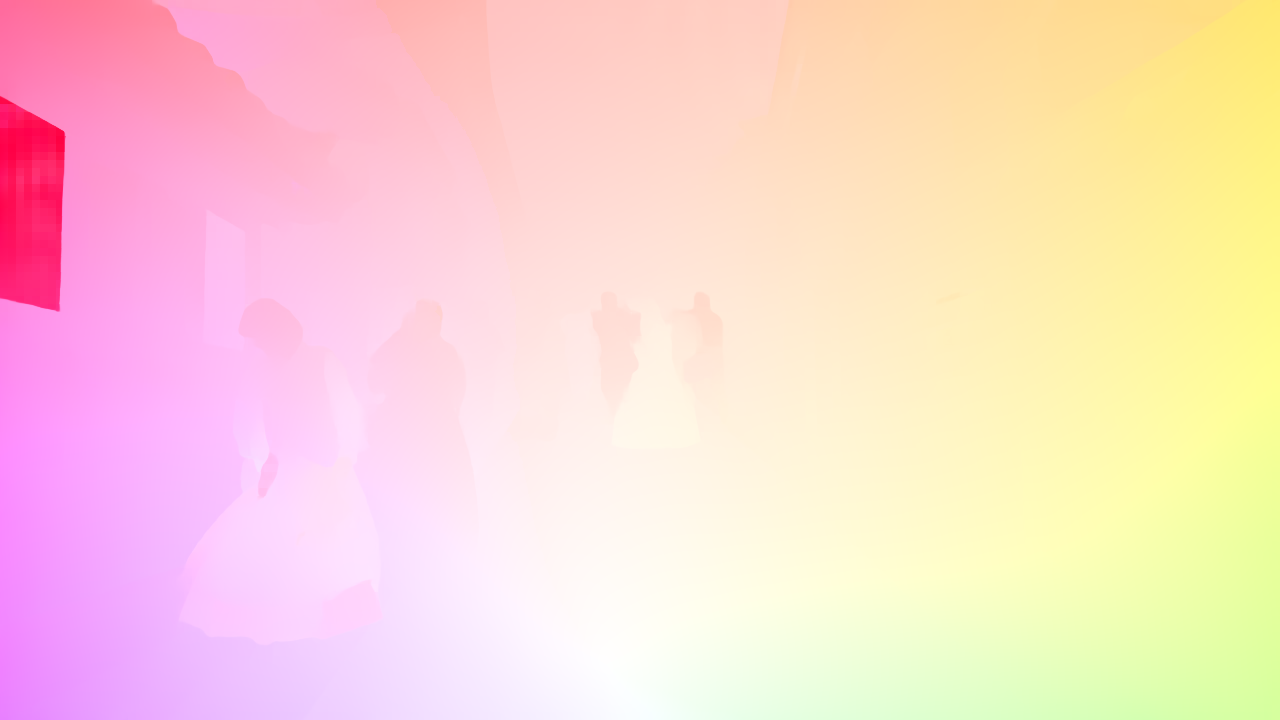} & \includegraphics[width=0.25\textwidth]{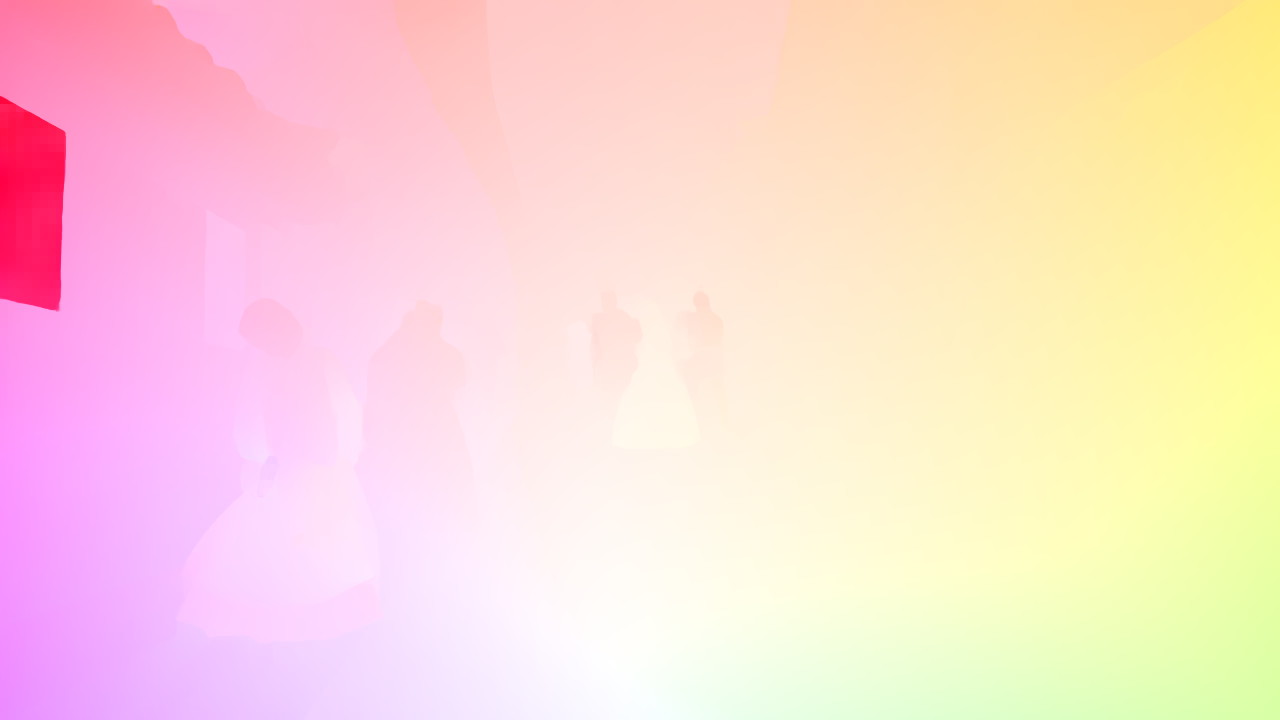}\\

        \includegraphics[width=0.25\textwidth]{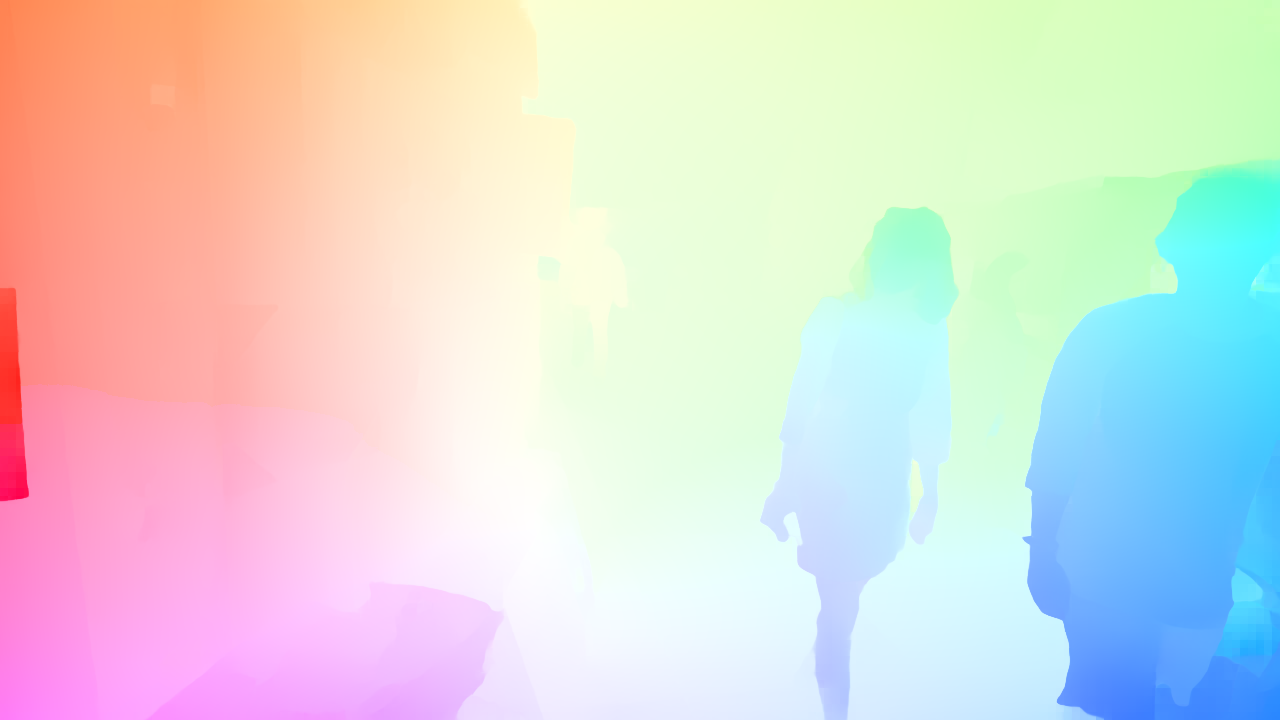} &
        \includegraphics[width=0.25\textwidth]{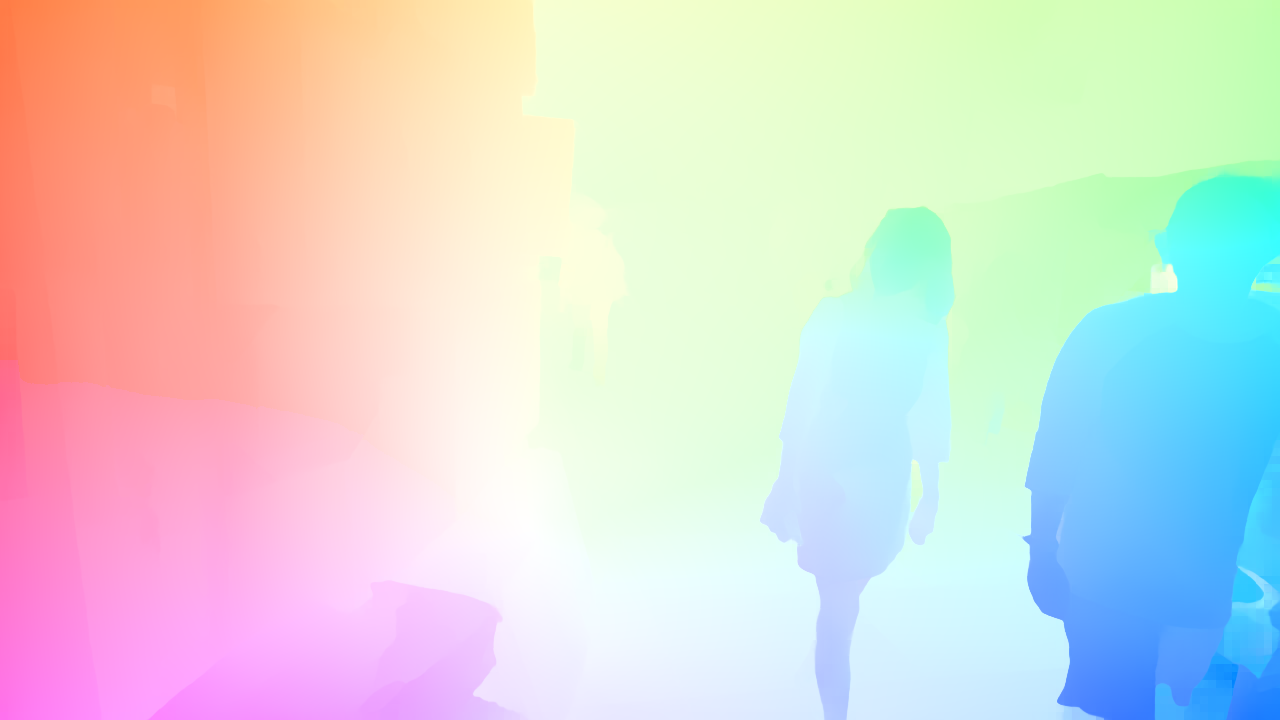} & \includegraphics[width=0.25\textwidth]{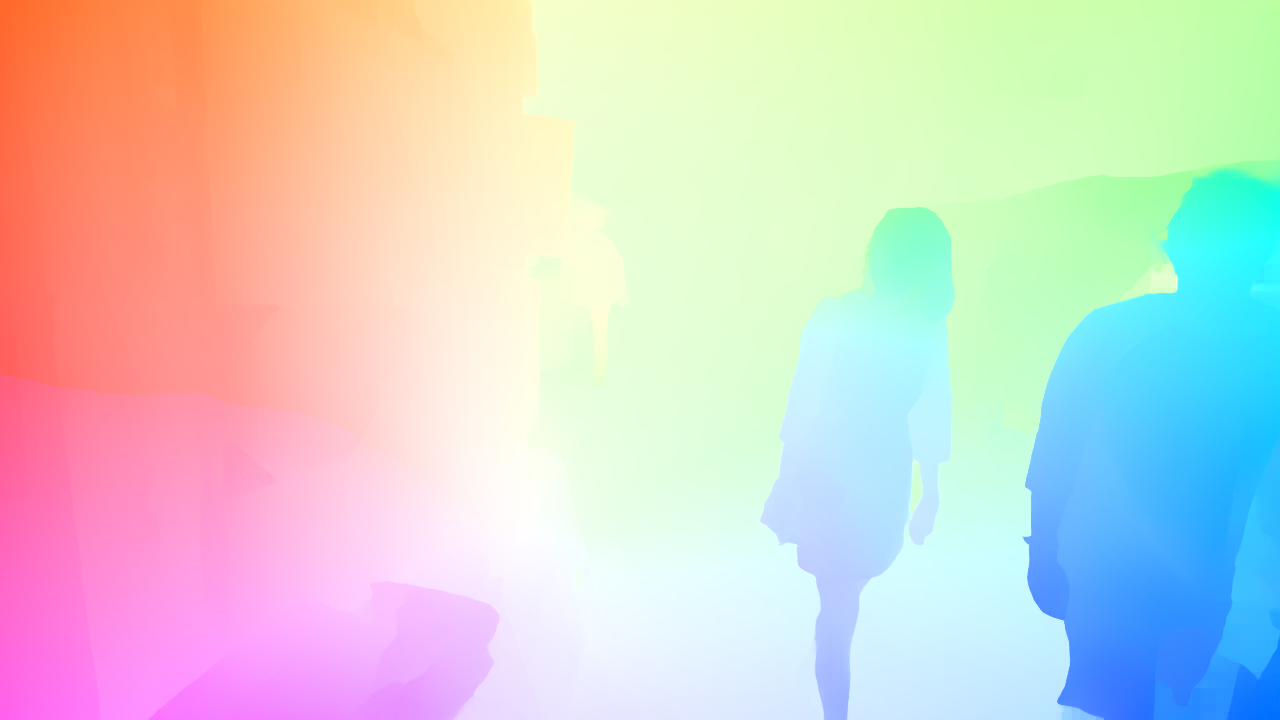} & \includegraphics[width=0.25\textwidth]{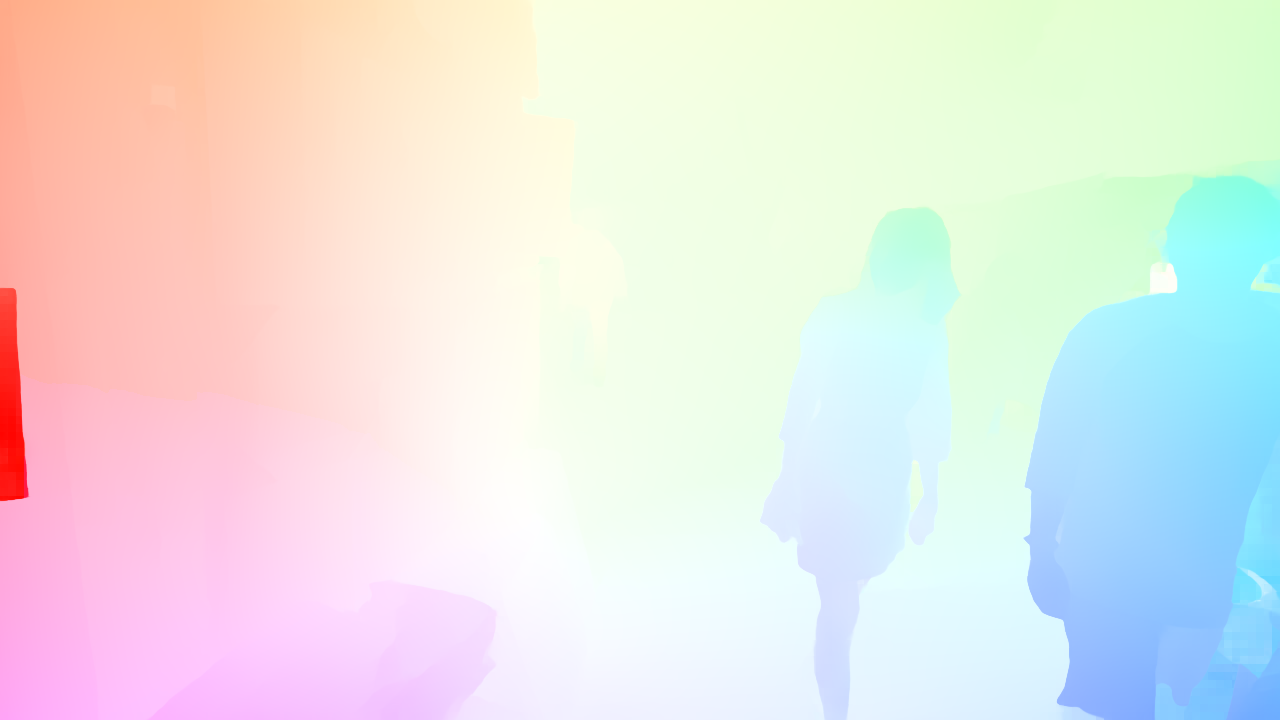} & \includegraphics[width=0.25\textwidth]{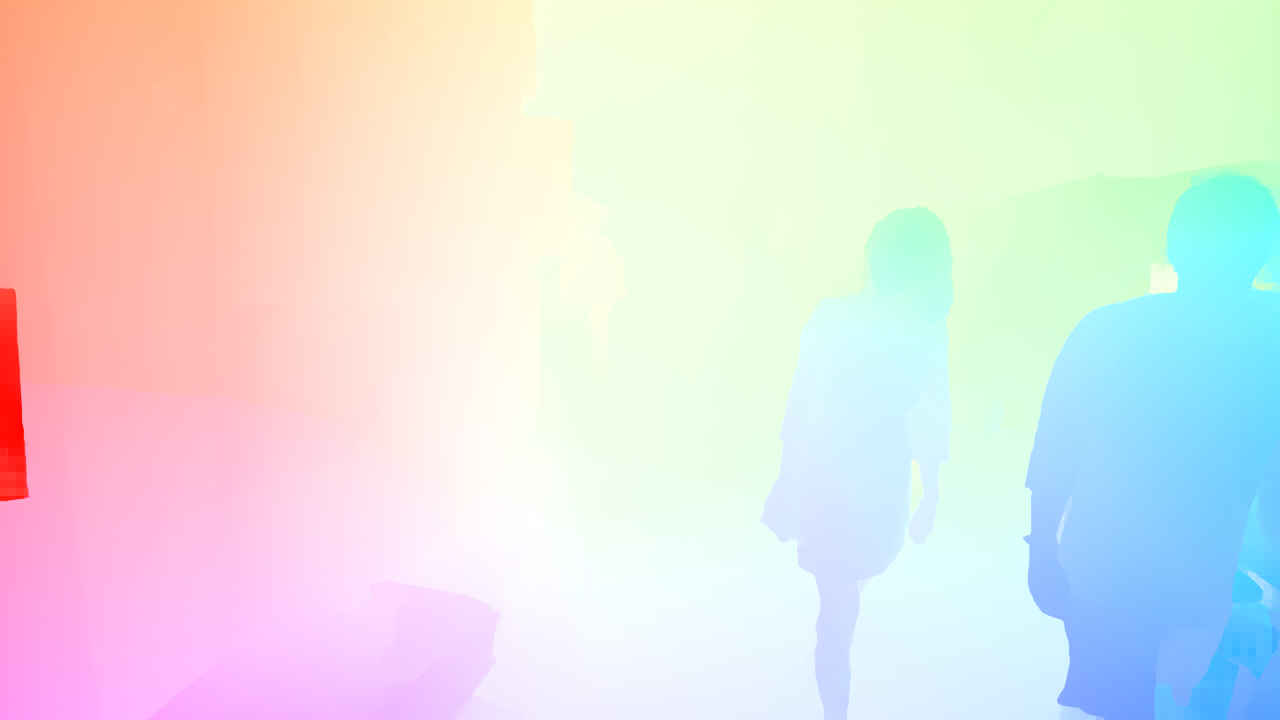}\\

    \end{tabular}}
    \caption{Comparison of optical flow (visualized) computed using RAFT on different state-of-the-art methods. Note that the hue represents the flow direction, while the saturation represents the flow magnitude. The optical flow computed on StableVSR results is more similar to the reference flow than the other methods. Results on sequences 000, 011 and 020 of REDS4, respectively.}
    \label{fig:ofcomparison}
\end{figure}

\minisection{Temporal consistency results.} Both temporal consistency metrics in Table~\ref{tab:results} show StableVSR achieves more temporally-consistent results. We provide some demo videos as supplementary material to qualitatively assess this aspect. In Figure~\ref{fig:temporalprofiles}, we show a comparison among temporal profiles of RealBasicVSR~\cite{chan2022investigating}, which is the second-best method on REDS4~\cite{nah2019ntire} according to tLP~\cite{chu2020learning} in Table~\ref{tab:results}, and the proposed StableVSR. We also report the temporal profiles of SD$\times4$Upscaler, which represents the baseline model used by our method. The temporal profiles of StableVSR are more regular and consistent with the reference profiles compared to the other methods, reflecting better consistency. In \Cref{fig:ofcomparison}, we compare the optical flow computed on consecutive frames obtained from RVRT~\cite{liang2022recurrent}, which represents the second-best method on REDS4~\cite{nah2019ntire} according to tOF~\cite{chu2020learning} in Table~\ref{tab:results}, and the proposed StableVSR. We also report SD$\times4$Upscaler and RealBasicVSR~\cite{chan2022investigating} results. We can observe the proposed StableVSR allows obtaining an optical flow more similar to the reference flow than the other methods. RealBasicVSR~\cite{chan2022investigating} obtains second-best and worst results on REDS4~\cite{nah2019ntire} according to tLP~\cite{chu2020learning} and tOF~\cite{chu2020learning}, respectively. Instead, the proposed StableVSR obtains best performance according to both the metrics.

\subsection{Ablation study}
\label{subsec:ablation}

\begin{figure}[t]
    \centering
    \setlength{\tabcolsep}{1pt}
    \renewcommand{\arraystretch}{0.5} 
    \begin{minipage}[t]{0.48\textwidth}
    \adjustbox{width=\textwidth}{
    \begin{tabular}{ccccc}
    & \makecell{No guidance\\on $\tilde{x}_0$} &  \makecell{No motion\\compensation} & \makecell{No\\ Latent$\rightarrow$RGB\\ conversion} & Proposed\\
    \raisebox{.02\height}{\rotatebox{90}{ \makecell{Conditioning\\frame $t=200$}}} & 
    \includegraphics[width=0.4\textwidth]{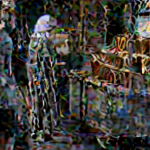} &
    \includegraphics[width=0.4\textwidth]{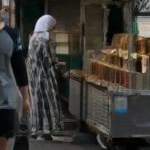} & \includegraphics[width=0.4\textwidth]{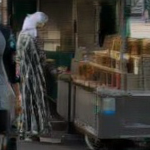} & \includegraphics[width=0.4\textwidth]{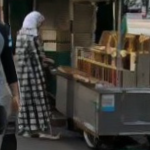}\\
    
    \raisebox{.04\height}{\rotatebox{90}{ \makecell{Conditioning\\frame $t=50$}}} & 
    \includegraphics[width=0.4\textwidth]{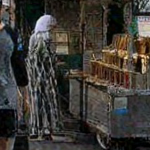} &
    \includegraphics[width=0.4\textwidth]{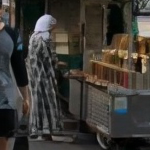} & \includegraphics[width=0.4\textwidth]{images/main/Ablation/Approximated_x0/Noconversion/181_011.png} &  \includegraphics[width=0.4\textwidth]{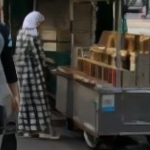}\\
    
    \raisebox{.35\height}{\rotatebox{90}{Frame $10$}} & 
    \includegraphics[width=0.4\textwidth]{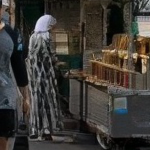} & \includegraphics[width=0.4\textwidth]{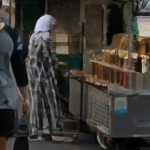} & \includegraphics[width=0.4\textwidth]{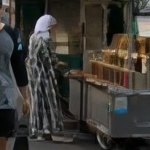} & \includegraphics[width=0.4\textwidth]{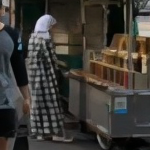}\\
    
    \raisebox{.35\height}{\rotatebox{90}{Frame $11$}} & 
    \includegraphics[width=0.4\textwidth]{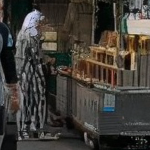} & \includegraphics[width=0.4\textwidth]
    {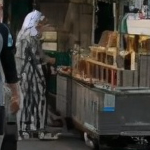} & \includegraphics[width=0.4\textwidth]{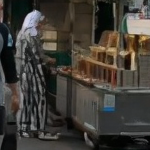} & \includegraphics[width=0.4\textwidth]{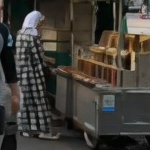}\\
    \end{tabular}}
    \subcaption{Ablation experiments on Temporal Texture Guidance. Using $x_t$ propagates noise. When motion compensation is not used, fine-detail information cannot be correctly used. Applying motion compensation in the latent space leads to undesired artifacts in the guidance. The proposed guidance solves these problems. 
    }
    \label{fig:x0options}
    \end{minipage}
    \hfill
    \begin{minipage}[t]{0.5\textwidth}
    \adjustbox{width=\textwidth}{
    \begin{tabular}{ccccc}
    & Single-frame & Auto-regressive & Proposed \\
        
    \raisebox{.5\height}{\rotatebox{90}{Frame 0}} & \includegraphics[width=0.4\textwidth]{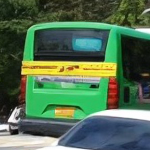} & \includegraphics[width=0.4\textwidth]{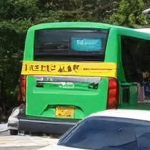} & \includegraphics[width=0.4\textwidth]{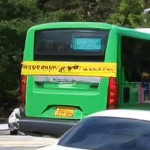}\\
    
    \raisebox{0.35\height}{\rotatebox{90}{Frame 53}} & \includegraphics[width=0.4\textwidth]{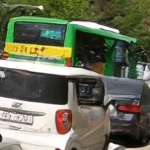} & \includegraphics[width=0.4\textwidth]{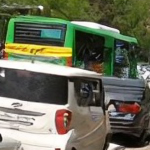} & \includegraphics[width=0.4\textwidth]{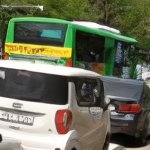} \\
    
    \raisebox{.35\height}{\rotatebox{90}{Frame 54}} & \includegraphics[width=0.4\textwidth]{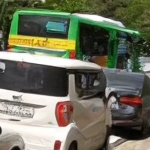} & \includegraphics[width=0.4\textwidth]{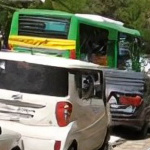} & \includegraphics[width=0.4\textwidth]{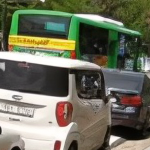} \\
    \end{tabular}}
    \subcaption{Ablation experiments on Frame-wise Bidirectional Sampling strategy. Single-frame sampling introduces temporal inconsistency. Auto-regressive sampling shows the error accumulation problem. The proposed sampling solves both problems.}
    \label{fig:samplingstrategycomparison}
    \end{minipage}
    \caption{Ablation experiments, qualitative results. For \quotes{No guidance on $\tilde{x}_0$} experiment, we use guidance on $x_t$. For \quotes{No Latent$\rightarrow$RGB conversion} experiment, the aligned latent is converted to RGB just for visualization.}
\end{figure}

\minisection{Temporal Texture Guidance.}
We evaluate the effectiveness of the Temporal Texture Guidance design by removing one of the operations involved in its computation.
Quantitative and qualitative results are shown in Table~\ref{tab:ablation} (upper part) and Figure~\ref{fig:x0options}, respectively.
Using guidance on $x_t$ instead of $\tilde{x}_0$ leads to very noisy frames. These noisy frames cannot provide adequate information when $t$ is far from 0.
With no motion compensation, the spatial information is not aligned with respect to the current frame and cannot be properly used. Applying motion compensation in the latent space introduces distortions in the guidance, as also shown in Figure~\ref{fig:motioncompensation}.
In all these cases, temporal consistency at fine-detail level cannot be achieved. The proposed approach provides detail-rich and spatially-aligned texture guidance at every sampling step $t$, leading to better temporal consistency. Additional results are reported in the supplementary material.

\begin{table}[t]
    \centering
    \caption{Ablation experiments, quantitative results. Perceptual metrics are marked with $\star$, reconstruction metrics with $\diamond$, and temporal consistency metrics with $\bullet$. 
    Best results in bold text. For \quotes{No guidance on $\tilde{x}_0$} experiment, we use guidance on $x_t$. In these experiments, the proposed solution achieves better results in terms of frame quality and temporal consistency. Results computed on center crops of $512\times512$ resolution of REDS4.}
    \label{tab:ablation}
    \adjustbox{width=\textwidth}{%
        \begin{tabular}{cccccccc}
                \toprule
               Ablated component & Experiment name & tLP$\bullet$$\downarrow$ & tOF$\bullet$$\downarrow$ &
                LPIPS$\star$$\downarrow$ & 
                DISTS$\star$$\downarrow$ & 
                PSNR$\diamond$$\uparrow$ & 
                SSIM$\diamond$$\uparrow$
                \\\midrule

                \multirow{4}{*}{\makecell{Temporal \\Texture\\ Guidance}} & No guidance on $\tilde{x}_0$ & 38.16 & 3.34 & 0.132 & 0.094 & 24.74 & 0.698 \\
                &No motion comp. & 18.97 & 3.47 & 0.116 & 0.077 & 25.70 & 0.749 \\
                &No $\text{Latent}\rightarrow\text{RGB}$ conv. & 21.17 & 3.32 & 0.113 & 0.076 & 25.78 & 0.752\\
                &Proposed & \textbf{6.16} & \textbf{2.84} & \textbf{0.095} & \textbf{0.067} & \textbf{27.14} & \textbf{0.799}\\\midrule
                \multirow{4}{*}{\makecell{Frame-wise\\Bidirectional\\Sampling}}&Single-frame & 14.67 & 3.99 & 0.121 & 0.087 & 25.49 & 0.729\\
                &Auto-regressive & 8.61 & 3.39 & 0.120 & 0.082 & 25.78 & 0.745\\
                &Unidirectional & 6.36 & 2.94 & 0.097 & 0.069 & 27.08 & 0.769 \\
                &Proposed & \textbf{6.16} & \textbf{2.84} & \textbf{0.095} & \textbf{0.067} & \textbf{27.14} & \textbf{0.799}\\
                \bottomrule
            \end{tabular}%
        }
\end{table}

\minisection{Frame-wise Bidirectional Sampling strategy.}
We compare the Frame-wise Bidirectional Sampling strategy with: single-frame sampling, \ie no temporal conditioning; auto-regressive sampling, \ie the previous upscaled frame is used as guidance for the current one; frame-wise unidirectional sampling, \ie only forward information propagation. The results are quantitatively and qualitatively evaluated in Table~\ref{tab:ablation} (bottom part) and Figure~\ref{fig:samplingstrategycomparison}, respectively.
Single-frame sampling leads to poor results and introduces temporal inconsistency due to the differences in the synthesized frame details. The auto-regressive approach has the problem of error accumulation, which is propagated to the next frames. Unidirectional sampling unbalances the information propagation, as only future frames receive information from the past ones, limiting the overall performance. The proposed Frame-wise Bidirectional Sampling solves these problems, leading to better and more consistent results.

\section{Discussion and limitations}
\label{sec:discussion}

\minisection{Reconstruction quality results.}
We focus on using DMs to enhance the perceptual quality in VSR. Under limited model capacity, improving perceptual quality inevitably leads to a decrease in reconstruction quality~\cite{blau2018perception}.
Recent works on single image super-resolution using DMs~\cite{saharia2022image,li2022srdiff,gao2023implicit} reported lower reconstruction quality when compared to regression-based methods~\cite{chen2021learning,lim2017enhanced}. This is related to the high generative capability of DMs, which may generate some patterns that help improve perceptual quality but negatively affect reconstruction quality.
Although most VSR methods target reconstruction quality, various studies~\cite{liu2022video,rota2023video} highlight the urgent need to address perceptual quality. We take a step in this direction. We believe improving perceptual or reconstruction quality is a matter of choice: for some application areas like the military, reconstruction error is more important, but for many areas like the film industry, gaming, and online advertising, perceptual quality is key.

\minisection{Model complexity.}
The overall number of model parameters in StableVSR is about $\times35$ higher than the compared methods, with a consequent increase in inference time and memory requirements. The iterative refinement process of DMs inevitably increases inference time. StableVSR takes about 100 seconds to upscale a video frame to a $1280\times720$ target resolution on an NVIDIA Quadro RTX 6000 using 50 sampling steps. In future works, we plan to incorporate current research in speeding up DMs~\cite{zheng2023fast, liu2023instaflow}, which allows reducing the number of sampling steps and decreasing inference time.

\section{Conclusion}
\label{sec:conclusion}
We proposed StableVSR, a method for VSR based on DMs that enhances the perceptual quality while ensuring temporal consistency through the synthesis of realistic and temporally-consistent details. We introduced the Temporal Conditioning Module into a pre-trained DM for SISR to turn it into a VSR method. TCM uses the Temporal Texture Guidance with spatially-aligned and detail-rich texture information from adjacent frames to guide the generative process of the current frame toward the generation of high-quality results and ensure temporal consistency. At inference time, we introduced the Frame-wise Bidirectional Sampling strategy to better exploit temporal information, further improving perceptual quality and temporal consistency. We showed in a comparison with state-of-the-art methods for VSR that StableVSR better enhances the perceptual quality of upscaled frames while ensuring superior temporal consistency.

\minisection{Acknowledgments.} We acknowledge projects TED2021-132513B-I00 and PID2022-143257NB-I00, financed by MCIN/AEI/10.13039/501100011033 and FSE+ by the European Union NextGenerationEU/PRTR, and the Generalitat de Catalunya CERCA Program. This work was partially supported by the MUR under the grant “Dipartimenti di Eccellenza 2023-2027" of the Department of Informatics, Systems and Communication of the University of Milano-Bicocca, Italy.

%
%
\bibliographystyle{splncs04}
\bibliography{manuscript}

\title{Enhancing Perceptual Quality in Video Super-Resolution through Temporally-Consistent Detail Synthesis using Diffusion Models\\- Supplementary Material -} 

\author{}
\institute{}

\titlerunning{Stable Video Super-Resolution (StableVSR)}
\authorrunning{C.~Rota et al.}

\maketitle

\noindent
This supplementary file provides additional details that were not included in
the main paper due to page limitations. Demo videos are available on the project page\footnote{\url{https://github.com/claudiom4sir/StableVSR}}.

\section{Additional methodology details}

\subsection{Description of the pre-trained LDM for SISR}

The proposed StableVSR is built upon a pre-trained Latent Diffusion Model (LDM) for single image super-resolution (SISR). We use Stable Diffusion $\times4$ Upscaler (SD$\times4$Upscaler)\footnote{\url{https://huggingface.co/stabilityai/stable-diffusion-x4-upscaler}}. It follows the LDM framework~\cite{rombach2022high}, which performs the iterative refinement process into a latent space and uses the VAE decoder $\mathcal{D}$~\cite{esser2021taming} to decode latents into RGB images.
Starting from a low-resolution RGB image LR (conditioning image) and an initial noisy latent $x_T$, the denoising UNet $\epsilon_\theta$ is used to generate the high-resolution counterpart via an iterative refinement process. In this process, noise is progressively removed from $x_t$ guided by LR. After a defined number of sampling steps, the obtained latent $x_0$ is decoded using the VAE decoder $\mathcal{D}$~\cite{esser2021taming} into a high-resolution RGB image $\overline{\text{HR}}$. The obtained image $\overline{\text{HR}}$ has a $\times4$ higher resolution than the low-resolution image LR, as $\mathcal{D}$ performs $\times 4$ upscaling. In practice, the low-resolution RGB image LR and the initial noisy latent $x_T$ are concatenated along the channel dimension and inputted to the denoising UNet.

\subsection{Bidirectional information propagation in the Frame-wise Bidirectional Sampling strategy}

\begin{figure}[t]
    \centering
    \includegraphics[width=\textwidth]{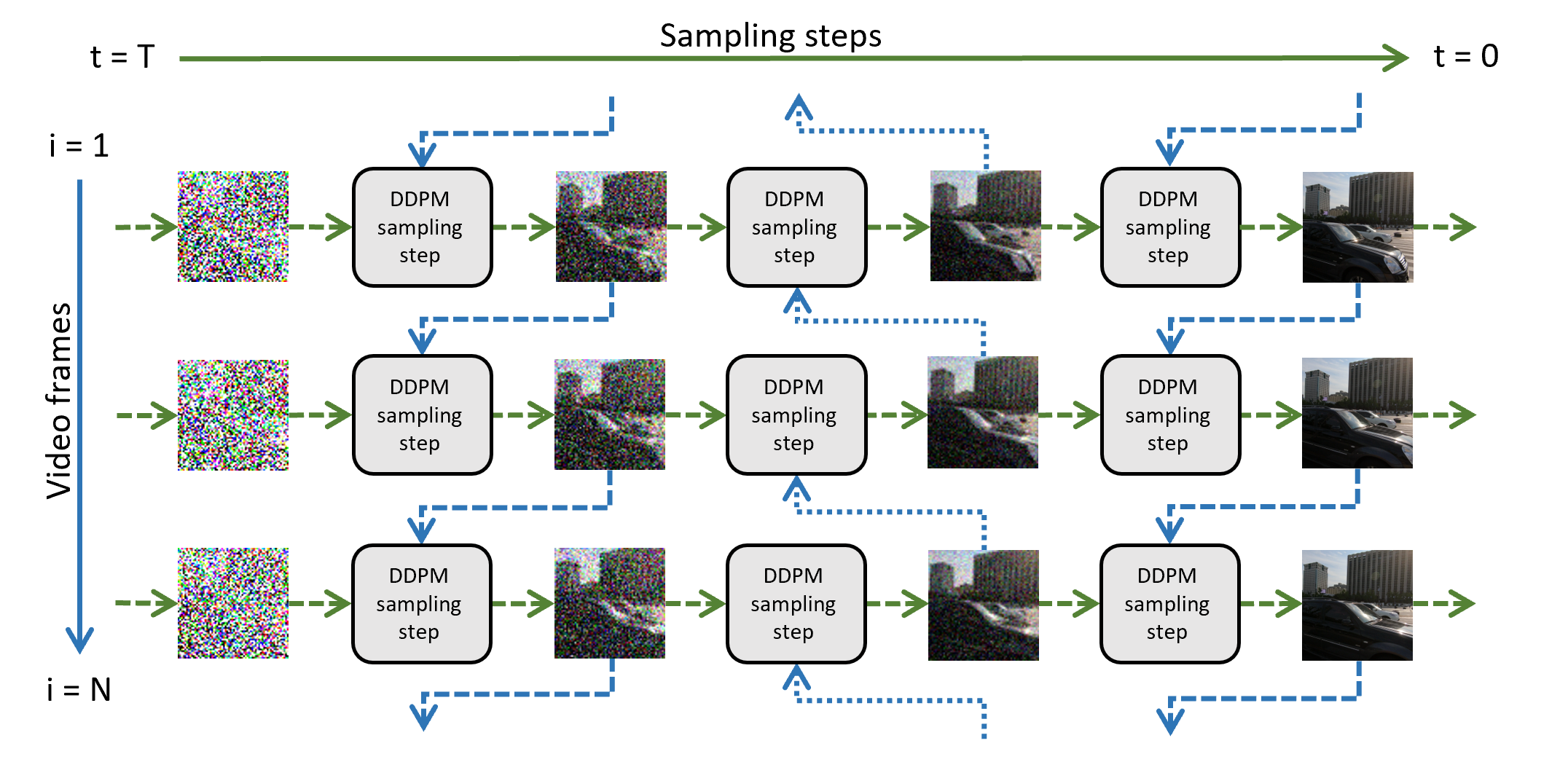}
    \caption{Graphical representation of the proposed Frame-wise Bidirectional Sampling strategy. The green flow propagates information forward in sampling time while the blue flow alternately propagates it forward and backward in video time. Forward propagation is shown with dashed lines, while backward propagation with dotted lines.}
    \label{fig:fwbss}
\end{figure}

We show in Figure~\ref{fig:fwbss} a graphical representation of the proposed Frame-wise Bidirectional Sampling strategy to better show the bidirectional information propagation.
We take a sampling step $t$ in video time $i=1, ..., N$ before moving to the next sampling step $t-1$. At every sampling step, we invert the video time order for processing: from $i=1, ..., N$ to $i=N, ..., 1$. For the generation of $x^i_{t-1}$, we start from $\tilde{x}^{i-1}_0$ and $x^i_{t}$. Since $\tilde{x}^{i-1}_0$ is related to the previous frame, it provides information from the past. In addition, since $x^i_{t}$ is generated starting from $\tilde{x}^{i+1}_{0}$ and $x^i_{t+1}$, it contains information from future frames, which is implicitly propagated to the current sampling step. As a consequence, $x^i_{t-1}$ benefits from past information from $\tilde{x}^{i-1}_0$ due to the forward direction of the current sampling step, and future information from $x^i_{t}$ due to the backward direction of the previous sampling step.

\section{Additional experiments}

\subsection{Architecture details}
We report the StableVSR architecture details in Table~\ref{tab:implementationdetails+}. 
\begin{table}
    \caption{Architecture details of StableVSR.}
    \label{tab:implementationdetails+}
    \centering
    \begin{tabular}{cccc}
    \toprule
         & Denoising UNet & Temporal Conditioning Module & VAE decoder\\\midrule
         Downscaling & $\times8$& $\times8$&-\\
         Upscaling & $\times8$& -&$\times4$\\
         Input channels & 7 & 3 & 4\\
         Output channels & 4 & - & 3\\
         Trainable& No & Yes & No\\
         Parameters & 473 M & 207 M& 32 M\\
         \bottomrule
    \end{tabular}
\end{table}
We can identify three main components: denoising UNet, Temporal Conditioning Module (TCM), and VAE decoder~\cite{esser2021taming}.
Following ControlNet~\cite{zhang2023adding}, we freeze the weights of the denoising UNet during training. We only train TCM for video adaptation. 
We apply spatial guidance on the low-resolution frame via concatenation, \ie the noisy latent $x^i_t$ (4 channels) is directly concatenated with the low-resolution frame $\text{LR}^i$ (3 channels) along the channel dimension. The temporal guidance is instead provided via TCM, which receives Temporal Texture Guidance $\widetilde{\text{HR}}^{i-1\rightarrow i}$ as input (3 channels). Once the iterative refinement process is complete, the VAE decoder $\mathcal{D}$~\cite{esser2021taming} receives the final latent of a frame $i$ as input, \ie $x^i_0$, and converts it into an RGB frame. This latent-to-RGB conversion applies $\times4$ upscaling, hence the output of the decoder represents the upscaled frame. The overall number of parameters in StableVSR (including the VAE decoder~\cite{esser2021taming}) is about 712 million.

\subsection{Additional comparison with state-of-the-art methods}

As in the main paper, we compare the proposed StableVSR with ToFlow~\cite{xue2019video}, EDVR~\cite{wang2019edvr},  TDAN~\cite{tian2020tdan},
MuCAN~\cite{li2020mucan},
BasicVSR~\cite{chan2021basicvsr}, BasicVSR$++$~\cite{chan2022basicvsr++}, RVRT~\cite{liang2022recurrent}, and RealBasicVSR~\cite{chan2022investigating}.

\begin{figure}
    \centering
    \begin{subfigure}[b]{\textwidth}
    \adjustbox{width=\textwidth}{
        \begin{tabular}{cccccc}
           Reference frame & Bicubic & BasicVSR++ & RVRT & StableVSR (ours) & Reference\\
           \includegraphics[height=0.2\textwidth]{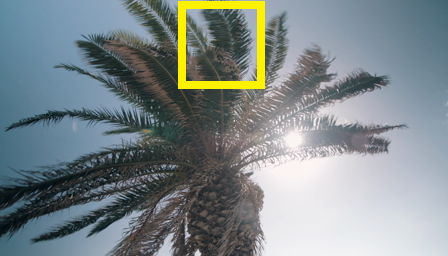}&

           \includegraphics[height=0.2\textwidth]{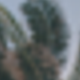}&

           \includegraphics[height=0.2\textwidth]{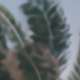} &

           \includegraphics[height=0.2\textwidth]{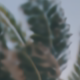}&

            \includegraphics[height=0.2\textwidth]{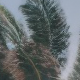} & 
            
            \includegraphics[height=0.2\textwidth]{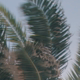}\\

            \includegraphics[height=0.2\textwidth]{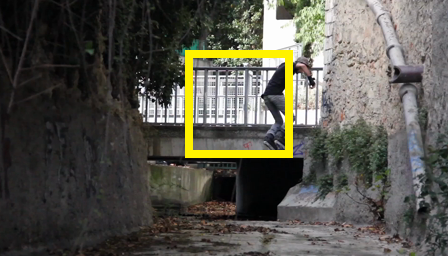}&

           \includegraphics[height=0.2\textwidth]{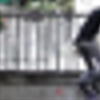}&

            \includegraphics[height=0.2\textwidth]{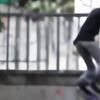} &

           \includegraphics[height=0.2\textwidth]{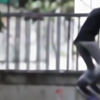}&

            \includegraphics[height=0.2\textwidth]{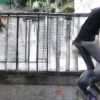} 
            
            &\includegraphics[height=0.2\textwidth]{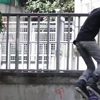}\\
            \includegraphics[height=0.2\textwidth]{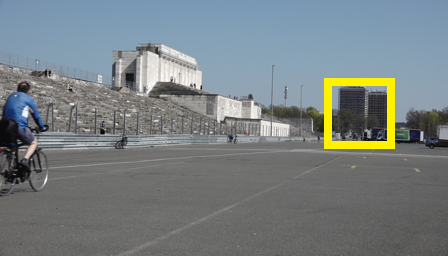}&

           \includegraphics[height=0.2\textwidth]{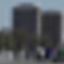}&

            \includegraphics[height=0.2\textwidth]{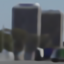} &

           \includegraphics[height=0.2\textwidth]{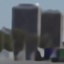}&

            \includegraphics[height=0.2\textwidth]{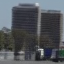} & 
            
            \includegraphics[height=0.2\textwidth]{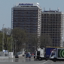}\\

            \includegraphics[height=0.2\textwidth]{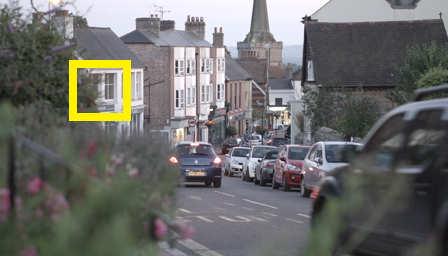}&

           \includegraphics[height=0.2\textwidth]{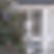}&

            \includegraphics[height=0.2\textwidth]{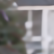} &

           \includegraphics[height=0.2\textwidth]{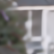}&

            \includegraphics[height=0.2\textwidth]{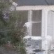} &
            \includegraphics[height=0.2\textwidth]{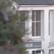}\\

            \includegraphics[height=0.2\textwidth]{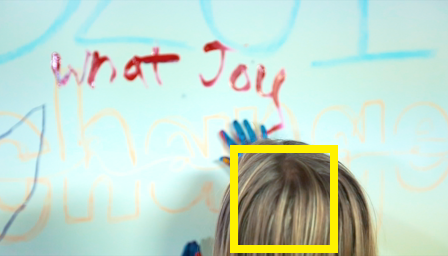}&

           \includegraphics[height=0.2\textwidth]{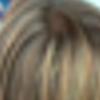}&
           
            \includegraphics[height=0.2\textwidth]{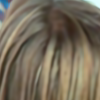} &

           \includegraphics[height=0.2\textwidth]{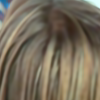}&

            \includegraphics[height=0.2\textwidth]{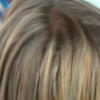} &  \includegraphics[height=0.2\textwidth]{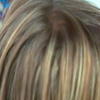}\\

            \includegraphics[height=0.2\textwidth]{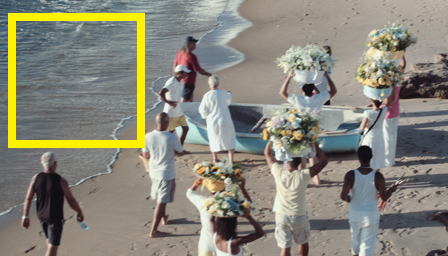}&

           \includegraphics[height=0.2\textwidth]{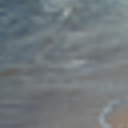}&

            \includegraphics[height=0.2\textwidth]{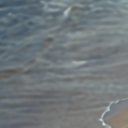} &

           \includegraphics[height=0.2\textwidth]{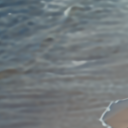}&

            \includegraphics[height=0.2\textwidth]{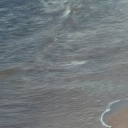} 
            &
            \includegraphics[height=0.2\textwidth]{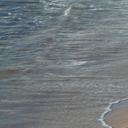}\\

            \includegraphics[height=0.2\textwidth]{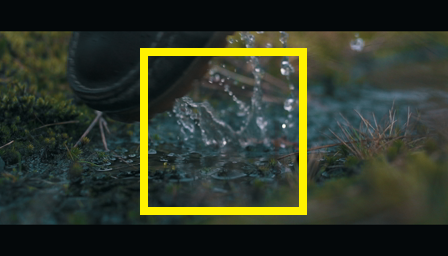}&

           \includegraphics[height=0.2\textwidth]{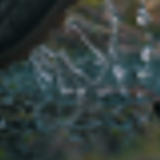}&

            \includegraphics[height=0.2\textwidth]{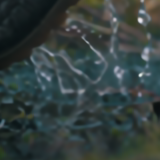} &

           \includegraphics[height=0.2\textwidth]{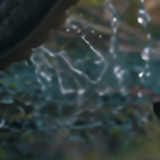}&

            \includegraphics[height=0.2\textwidth]{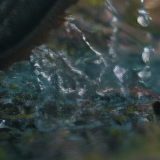} &\includegraphics[height=0.2\textwidth]{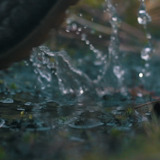}\\
           
        \end{tabular}}

        \caption{Results on Vimeo-90K-T.}
        \label{subfig:qualitativevimeo+}
        \end{subfigure}
        
        \vspace{0.5cm}
        \begin{subfigure}[b]{\textwidth}
        \adjustbox{width=\textwidth}{
        \begin{tabular}{cccccc}
          Reference frame & Bicubic & RVRT & RealBasicVSR & StableVSR (ours) & Reference\\

        \includegraphics[height=0.2\textwidth]{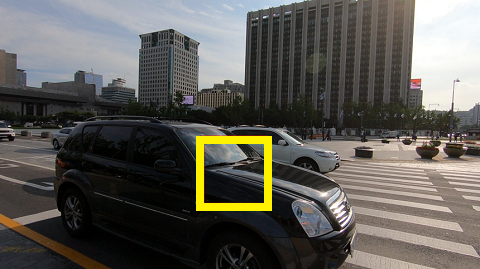} &
        \includegraphics[height=0.2\textwidth]{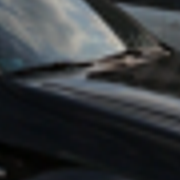} &
         \includegraphics[height=0.2\textwidth]{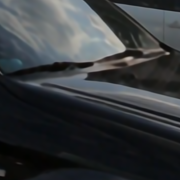} &
        \includegraphics[height=0.2\textwidth]{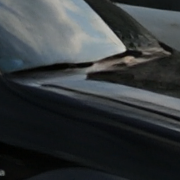} &
        \includegraphics[height=0.2\textwidth]{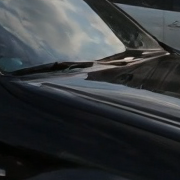} &
        \includegraphics[height=0.2\textwidth]{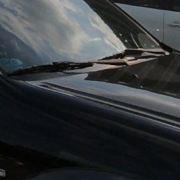}\\
        
        \includegraphics[height=0.2\textwidth]{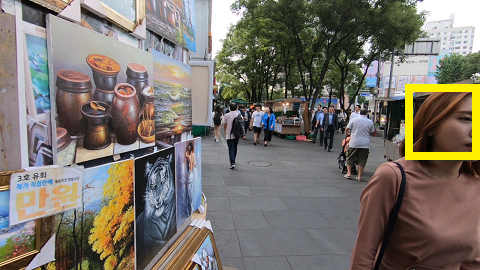} &
        \includegraphics[height=0.2\textwidth]{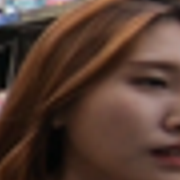} &
         \includegraphics[height=0.2\textwidth]{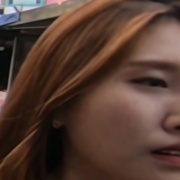} &
        \includegraphics[height=0.2\textwidth]{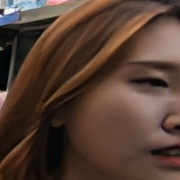} &
        \includegraphics[height=0.2\textwidth]{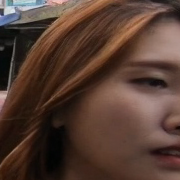} &
        \includegraphics[height=0.2\textwidth]{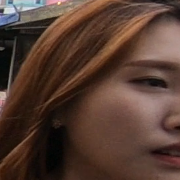}
    \end{tabular}
    }
    \caption{Results on REDS4.}
    \label{subfig:qualitativeredso+}
        \end{subfigure}
    \caption{Additional qualitative comparison with state-of-the-art methods for VSR. Only the proposed StableVSR correctly upscales complex textures.}
    \label{fig:qualitative+}   
\end{figure}

\minisection{Frame quality results.}
We report additional results using no-reference perceptual quality metrics, including MUSIQ~\cite{ke2021musiq}, CLIP-IQA~\cite{wang2023exploring} and NIQE~\cite{mittal2012making}. The results are reported in Table~\ref{tab:results+}. All the metrics highlight the proposed StableVSR achieves superior perceptual quality. The only exception is NIQE~\cite{mittal2012making} on REDS4~\cite{nah2019ntire}, which indicates StableVSR achieves the second-best results.
We show in Figure~\ref{fig:qualitative+} an additional qualitative comparison with BasicVSR$++$~\cite{chan2022basicvsr++} and RVRT~\cite{liang2022recurrent} on Vimeo-90K-T~\cite{xue2019video} (Figure~\ref{subfig:qualitativevimeo+}) and with RVRT~\cite{liang2022recurrent} and RealBasicVSR++~\cite{chan2022investigating} on REDS4~\cite{nah2019ntire} (Figure~\ref{subfig:qualitativeredso+}). We can observe the proposed StableVSR is the only method that correctly upscales complex textures while the other methods fail, producing blurred results.

\begin{table}[t]
\caption{Additional quantitative comparison with state-of-art methods for VSR using no-reference perceptual metrics. Best results in bold text. Almost all the metrics highlight the proposed StableVSR achieves better perceptual quality.}
\label{tab:results+}
\centering
\adjustbox{width=0.8\textwidth}{%
\begin{tabular}{ccccccc}
\toprule
                 \multirow{2}{*}{Method}& \multicolumn{3}{c}{Vimeo-90K-T}                         & \multicolumn{3}{c}{REDS4}\\\cmidrule(lr){2-4}\cmidrule(lr){5-7}
 &
  MUSIQ$\uparrow$ &
  CLIP-IQA$\uparrow$ &
  NIQE$\downarrow$ &
  MUSIQ$\uparrow$ &
  CLIP-IQA$\uparrow$ &
  NIQE$\downarrow$\\\midrule
Bicubic &
 23.27 & 0.358 & 8.44 & 26.89 & 0.304 & 6.85 \\
ToFlow   &   40.79    &   0.364    & 8.05 & - & - & - \\
EDVR  & -   &  -     &   -   &  65.44 & 0.367 & 4.15 \\
TDAN & 46.54    &  0.386     &  7.34 &   - & - & -     \\
MuCAN & 49.84 & 0.379 & 7.22 & 64.85 & 0.362 & 4.30 \\
BasicVSR   & 48.97 & 0.376 & 7.27 & 65.74 & 0.371 & 4.06 \\
BasicVSR$++$ & 50.11 & 0.383 & 7.12 & 67.00 & 0.381 & 3.87\\
RVRT   & 50.45 & 0.387 & 7.12 & 67.44 & 0.392 & 3.78 \\
RealBasicVSR & - & - & - & 67.03 & 0.374 & \textbf{2.53} \\
StableVSR (ours) & \textbf{50.97} & \textbf{0.414} & \textbf{5.99}  & \textbf{67.54} & \textbf{0.417} & 2.73\\\bottomrule

\end{tabular}%
}
\end{table}

\minisection{Temporal consistency results.} We can qualitatively assess the temporal consistency aspect of the proposed StableVSR in the demo videos. We compare StableVSR with SD$\times 4$Upscaler, which represents the baseline model used by StableVSR, and RealBasicVSR~\cite{chan2022investigating}, which represents the second-best method on REDS4~\cite{nah2019ntire} in terms of temporal consistency.

\subsection{Comparison with the DM video baseline}

We compare the proposed StableVSR with a DM video baseline containing 3D convolutions and temporal attention. 
Starting from the same pre-trained DM for SISR we use in StableVSR, i.e. SD$\times4$Upscaler, we implement the video baseline by introducing a temporal layer (3D convolutions + temporal attention) after each pre-trained spatial layer, as done in previous video generation methods~\cite{ho2022video,blattmann2023align,wu2023tune}. For training, we set the temporal window size to 5 consecutive frames and use the same training settings as in StableVSR. The only difference is the batch size, which is set to 8 instead of 32 due to memory constraints. We freeze the spatial layers and only train the temporal layers. 
Table~\ref{tab:comparisonbaseline} reports the results, where we can see the proposed StableVSR achieves better performance in both frame quality and temporal consistency. We attribute the lower performance of the DM video baseline to the limited temporal view, the inability to capture fine-detail image information, and the lack of proper frame alignment. StableVSR does not suffer from these problems, achieving better results.

\begin{table}[t]
\centering
\caption{Comparison with the DM video baseline. Perceptual metrics are marked with $\star$, reconstruction metrics with $\diamond$, and temporal consistency metrics with $\bullet$. Best results in bold text. The proposed StableVSR achieves better results in terms of frame quality and temporal consistency. Results computed on center crops of $512\times512$ resolution of REDS4.}
\label{tab:comparisonbaseline}
            \begin{tabular}{ccccccc}
                \toprule
                Method& tLP$\bullet$$\downarrow$ & tOF$\bullet$$\downarrow$ &
                LPIPS$\star$$\downarrow$ & 
                DISTS$\star$$\downarrow$ & 
                PSNR$\diamond$$\uparrow$ & 
                SSIM$\diamond$$\uparrow$
                \\\midrule
                Video baseline & 13.08 &2.92  & 0.113 & 0.075 & 26.27 & 0.771 \\
                StableVSR (ours) & \textbf{6.16} & \textbf{2.84} & \textbf{0.095} & \textbf{0.067} & \textbf{27.14} & \textbf{0.799}\\
                \bottomrule
            \end{tabular}
\end{table}

\subsection{Additional ablation study}
\begin{figure}[t]
    \centering
    \setlength{\tabcolsep}{1pt}
    \renewcommand{\arraystretch}{0.5} 
    \adjustbox{width=\textwidth}{
    \begin{tabular}{ccccc}
       & \makecell{No guidance\\on $\tilde{x}_0$} &  \makecell{No motion\\compensation} & \makecell{No\\ Latent$\rightarrow$RGB\\ conversion} & Proposed\\

        \raisebox{.45\height}{\rotatebox{90}{Frame $36$}} & \includegraphics[width=0.2\textwidth]{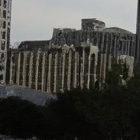} &
       \includegraphics[width=0.2\textwidth]{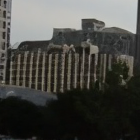} &
       \includegraphics[width=0.2\textwidth]{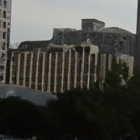} &
       \includegraphics[width=0.2\textwidth]{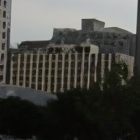}\\

       \raisebox{.45\height}{\rotatebox{90}{Frame $37$}} & \includegraphics[width=0.2\textwidth]{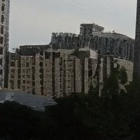} &
       \includegraphics[width=0.2\textwidth]{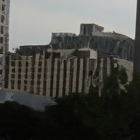} &
       \includegraphics[width=0.2\textwidth]{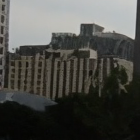} &
       \includegraphics[width=0.2\textwidth]{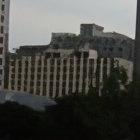} \\

       \raisebox{.45\height}{\rotatebox{90}{Frame $38$}} &\includegraphics[width=0.2\textwidth]{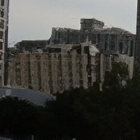} &
       \includegraphics[width=0.2\textwidth]{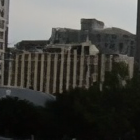} &
       \includegraphics[width=0.2\textwidth]{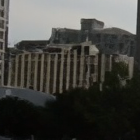} &
       \includegraphics[width=0.2\textwidth]{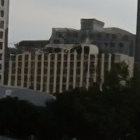}\\
       
    \end{tabular}
    }
    \caption{Additional ablation experiments for the Temporal Texture Guidance. We show the results obtained on three consecutive frames. Only the proposed solution ensures temporal consistency at the fine-detail level over time. Results on sequence 015 of REDS4.}
    \label{fig:ablationTTG+}
\end{figure}

\minisection{Temporal Texture Guidance.} 
In Figure~\ref{fig:ablationTTG+}, we provide additional results related to the ablation study on the Temporal Texture Guidance. We can observe that only the proposed design for the Temporal Texture Guidance ensures temporal consistency at the fine-detail level over time.

\subsection{Impact of sampling steps}
\label{subsec:additionalexp}

\begin{figure}[t]
    \centering
    \begin{subfigure}{0.19\textwidth}\includegraphics[height=\textwidth]{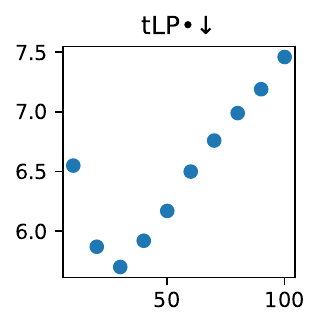}
    \end{subfigure}
    \begin{subfigure}{0.19\textwidth}\includegraphics[height=\textwidth]{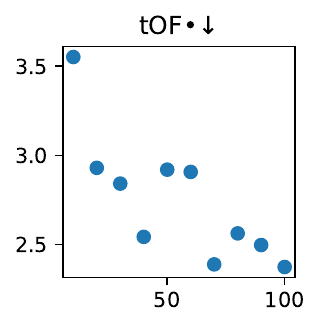}
    \end{subfigure}
    \begin{subfigure}{0.19\textwidth}\includegraphics[height=\textwidth]{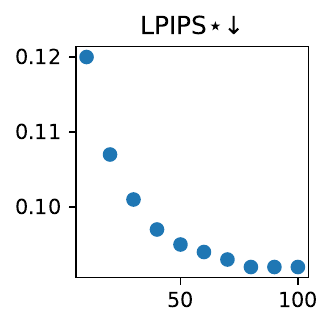}
    \end{subfigure}
    \begin{subfigure}{0.19\textwidth}\includegraphics[height=\textwidth]{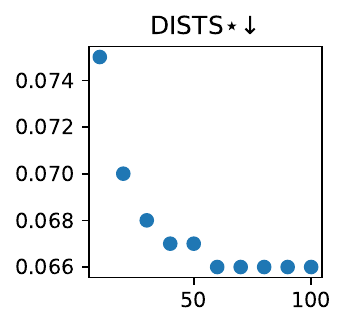}
    \end{subfigure}
    \begin{subfigure}{0.19\textwidth}\includegraphics[height=\textwidth]{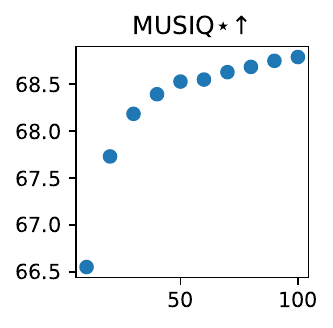}
    \end{subfigure}
    \begin{subfigure}{0.19\textwidth}\includegraphics[height=\textwidth]{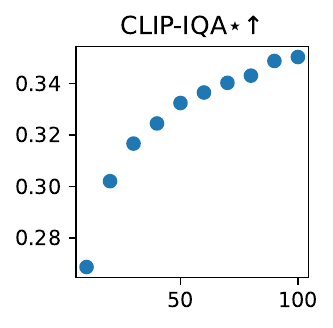}
    \end{subfigure}
    \begin{subfigure}{0.19\textwidth}\includegraphics[height=\textwidth]{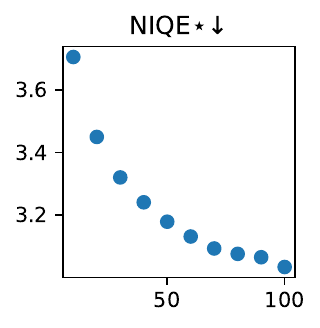}
    \end{subfigure}
    \begin{subfigure}{0.19\textwidth}\includegraphics[height=\textwidth]{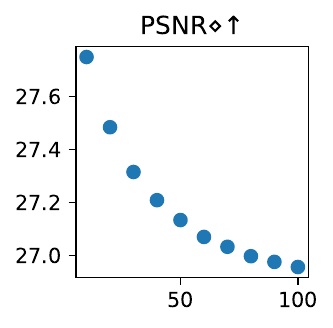}
    \end{subfigure}
    \begin{subfigure}{0.19\textwidth}\includegraphics[height=\textwidth]{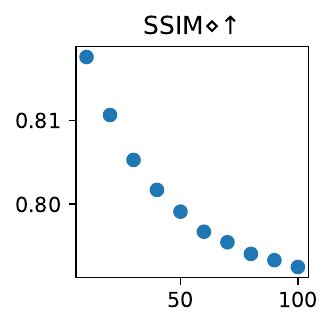}
    \end{subfigure}
    \caption{Performance changes as the number of sampling steps varies. The $x$ axis represents sampling steps, while the $y$ axis metric values. Perceptual metrics are marked with $\star$, reconstruction metrics with $\diamond$, and temporal consistency metrics with $\bullet$. Increasing the sampling steps improves perceptual quality while deteriorating reconstruction quality. Results computed on center crops of $512\times512$ 
    resolution of REDS4.}
    \label{fig:samplingsteps}
\end{figure}

We study how the performance changes as the number of sampling steps varies. Figure~\ref{fig:samplingsteps} shows the results obtained by increasing the number of sampling steps from 10 to 100. Reconstruction quality metrics, i.e. PSNR and SSIM~\cite{wang2004image}, deteriorate with more sampling steps. Conversely, perceptual quality metrics, i.e. LPIPS~\cite{zhang2018unreasonable}, DISTS~\cite{ding2020image}, MUSIQ~\cite{ke2021musiq}, CLIP-IQA~\cite{wang2023exploring}, NIQE~\cite{mittal2012making}, improve. We can attribute this behavior to the iterative refinement process of DMs, which progressively refines realistic image details that may not be perfectly aligned with the reference. We can observe the temporal consistency metric tLP~\cite{chu2020learning} reaches the best value using 30 steps, while tOF~\cite{chu2020learning} values are better as the number of sampling steps increases. According to these results, 50 sampling steps represent a good balance between perceptual quality and temporal consistency.

%
%

\end{document}